\documentclass[runningheads]{llncs}
\usepackage{graphicx}

\usepackage{tikz}
\usepackage{comment}
\usepackage{amsmath,amssymb} 
\usepackage{color}
\usepackage{xspace}
\usepackage{booktabs} 
\usepackage[implicit=false]{hyperref}

\usepackage[accsupp]{axessibility}  

\usepackage{color}
\usepackage{subcaption}
\usepackage{multirow}
\usepackage{wrapfig}
\newcommand{\search}{\textit{search*}{}}
\newcommand{\searchplus}{\textit{search}{}}
\newcommand{\optinit}{\textit{Opt-init}\xspace}
\newcommand{\optend}{\textit{Opt-end}\xspace}

\begin{document}
\pagestyle{headings}
\mainmatter
\def\ECCVSubNumber{4832}  

\title{AdvDO: Realistic Adversarial Attacks for Trajectory Prediction} 

\titlerunning{AdvDO}
%
\author{Yulong Cao\inst{1,2}\thanks{This work was done during an internship at NVIDIA} \and
Chaowei Xiao\inst{2} \and
Anima Anandkumar\inst{2,3}\and
Danfei Xu\inst{2}\and
Marco Pavone\inst{2,4}}
\authorrunning{Y. Cao et al.}
%
\institute{University of Michigan, Ann Arbor 
\and NVIDIA \and California Institute of Technology \and Stanford University}


\maketitle

\begin{abstract}
Trajectory prediction is essential for autonomous vehicles (AVs) to plan correct and safe driving behaviors. While many prior works aim to achieve higher prediction accuracy, few study the adversarial robustness of their methods. To bridge this gap, we propose to study the adversarial robustness of data-driven trajectory prediction systems. We devise an optimization-based adversarial attack framework that leverages a carefully-designed {\em differentiable dynamic model} to generate realistic adversarial trajectories.
Empirically, we benchmark the adversarial robustness of state-of-the-art prediction models and show that our attack increases
the prediction error for both general metrics and planning-aware metrics by more than 50\% and 37\%. We also show that our attack can lead an AV to drive off road or collide into other vehicles in simulation.
Finally, we demonstrate how to mitigate the adversarial attacks using an adversarial training scheme\footnote[1]{Our project website is at \href{https://robustav.github.io/RobustPred/}{https://robustav.github.io/RobustPred}}. 

\end{abstract}

\section{Introduction}

Trajectory forecasting is an integral part of modern autonomous vehicle (AV) systems. It allows an AV system to anticipate the future behaviors of other nearby road users and  plan its actions accordingly. 
Recent data-driven methods have shown remarkable performances on motion forecasting benchmarks~\cite{alahi2016social,ivanovic2019trajectron,salzmann2020trajectron++,yuan2021agent,rhinehart2018r2p2,rhinehart2019precog,kosaraju2019social}.
At the same time, for a safety-critical system like an AV, it is as essential for its components to be high-performing as it is for them to be reliable and robust. 
But few existing work have considered the robustness of these trajectory prediction models, especially when they are subject to deliberate adversarial attacks. 

\begin{figure}
    \centering
    \includegraphics[width=0.8\linewidth]{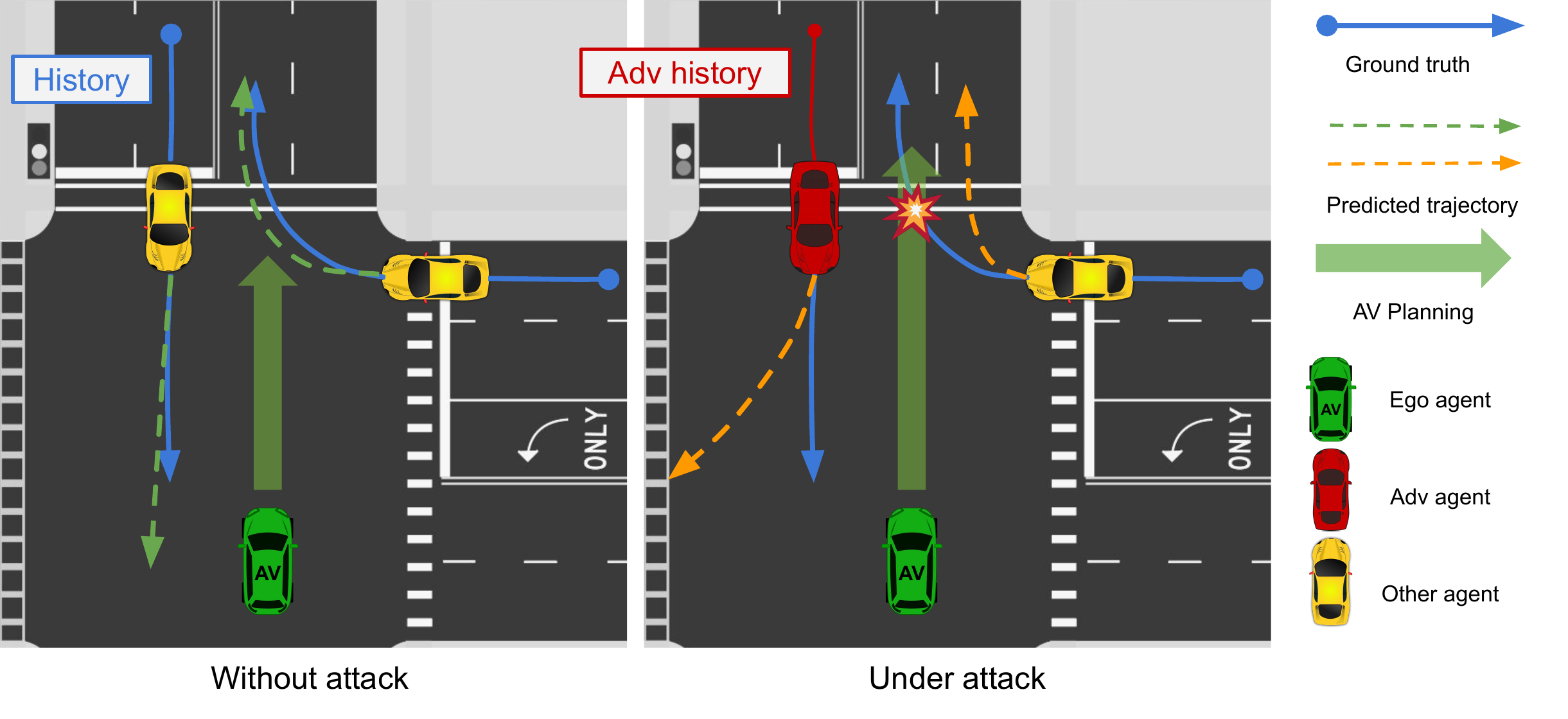}
    \caption{An example of attack scenarios on trajectory prediction. By driving along the crafted adversarial history trajectory, the adverial agent misleads the prediction of the AV systems for both itself and the other agent. As a consequence, the AV planning based on the wrong prediction results in a collision.}
    \label{fig:attack_overview}
\end{figure}

A typical adversarial attack framework consists of a threat model, i.e., a function that generates ``realistic'' adversarial samples, adversarial optimization objectives, and ways to systematically determine the influence of the attacks. However, a few key technical challenges remain in devising such a framework for attacking trajectory prediction models.

First, the threat model must synthesize adversarial trajectory samples that are 1) feasible subject to the physical constraints of the real vehicle (i.e. dynamically feasible), and 2) close to the nominal trajectories. The latter is especially important as a large alteration to the trajectory history conflates whether the change in future predictions is due to the vunerability of the prediction model or more fundamental changes to the meaning of the history.
To this front, we propose an attack method that uses a carefully designed {\em differentiable dynamic model} to generate adversarial trajectories that are both effective and realistic. Furthermore, through a gradient-based optimization process, we can generate adversarial trajectories efficiently and customize the adversarial optimization objectives to create different safety-critical scenarios. 

Second, not all trajectory prediction models react to attacks the same way. 
Features that are beneficial in benign settings may make a model more vulnerable to adversarial attacks. We consider two essential properties of modern prediction models: (1) motion property, which captures the influence of past agent states over future states; and (2) social property, which captures how the state of each agent affects others. Existing prediction models have proposed various architectures to explicitly models these properties either in silo~\cite{salzmann2020trajectron++} or jointly~\cite{yuan2021agent}.
Specifically, we design an attack framework that accounts for the above properties. 
We show that our novel attack framework can exploit these design choices. 
As illustrated in Figure~\ref{fig:attack_overview},  
by only manipulating the history trajectory of the adversarial agent, 
we are able to mislead the predicted future trajectory for the adversarial agent (i.e. incorrect prediction for left turning future trajectory of red car in Figure~\ref{fig:attack_overview}-right). 
Furthermore, we are able to mislead the prediction for {\em other} agent's behavior (i.e. turning right to turning left  for the yellow car in Figure~\ref{fig:attack_overview}-right). During the evaluation, we could evaluate these two goals respectively. It helps us fine-grained diagnose vulnerability of different models. 

Finally, existing prediction metrics such as average distance error (ADE) and final distance error (FDE) only measure errors of average cases and are thus too coarse for evaluating the effectiveness of adversarial attacks. They also ignore the influence of prediction errors in downstream planning and control pipelines in an AV stack.
To this end, we incorporate various metrics with semantic meanings such as \textit{off-road rates}, \textit{miss rates} and \textit{planning-aware metrics}~\cite{Ivan2022planpred} to systematically quantify the effectiveness of the attacks on prediction. 
We also conduct end-to-end attack on a prediction-planning pipeline by simulating the driving behavior of an AV in a close-loop manner. We demonstrate that the proposed attack can lead to both emergency brake and various of collisions of the AV. 

We benchmark the adversarial robustness of state-of-the-art trajectory prediction models~\cite{yuan2021agent,salzmann2020trajectron++} on the nuScenes dataset~\cite{caesar2020nuscenes}.
We show that our attack can increase prediction error by 50\% and 37\% on general metrics and planning-aware metrics, respectively. We also show that adversarial trajectories are realistic both quantitatively and qualitatively. Furthermore, we demonstrate that the proposed attack can lead to severe consequences in simulation.
Finally, we explore the mitigation methods with adversarial training using the proposed adversarial dynamic optimization method (AdvDO). We find that the model trained with the dynamic optimization increase the adversarial robustness by 54\%.

\section{Related works}
\label{sec:related_work}
\noindent\textbf{Trajectory Prediction.}
\label{subsec:related_work_trajectory_prediction}
Modern trajectory prediction models are usually deep neural networks that take state histories of agents as input and generate their plausible future trajectories. 
Accurately forecasting multiagent behaviors requires modeling two key properties:
(1) motion property, which captures the influence of past agent states over future states; (2) social property, which captures how the state of each agent affects others. Most prior works model the two properties separately~\cite{ivanovic2019trajectron,salzmann2020trajectron++,deo2021multimodal,suo2021trafficsim,kosaraju2019social}. For example, a representative method Trajactron++~\cite{salzmann2020trajectron++} summarizes temporal and inter-agent features using a time-sequence model and a graph network, respectively.
But modeling these two properties in silo ignores dependencies across time and agents. A recent work Agentformer~\cite{yuan2021agent} introduced a joint model 
that allows an agent’s state at one time to directly affect another agent’s state at a future time via a transformer model. 

At the same time, although these design choices for modeling motion and social properties may be beneficial in benign cases, they might affect a model's performance in unexpected ways when under adversarial attacks. Hence we select these two representative models~\cite{salzmann2020trajectron++,yuan2021agent} for empirical evaluation.


\noindent\textbf{Adversarial Traffic Scenarios Generation.} Adversarial traffic scenario generation is to synthesize traffic scenarios that could potentially pose safety risks\cite{ding2020learning,koren2019efficient,ding2021multimodal,abeysirigoonawardena2019generating,rempe2021strive}. Most prior approaches fall into two categories. The first aims to capture traffic scenarios distributions from real driving logs using generative models and sample adversarial cases from the distribution. For example, STRIVE~\cite{rempe2021strive} learns a latent generative model of traffic scenarios and then searches for latent codes that map to risky cases, such as imminent collisions. However, these latent codes may not correspond to real traffic scenarios. As shown in the supplementary materials, the method generates scenarios that are unlikely in the real world (e.g. driving on the wrong side of the road). Note that this is a fundamental limitation of generative methods, because almost all existing datasets only include safe scenarios, and it is hard to generate cases that are rare or non-existent in the data. 

Our method falls into the second category, which is to generate adversarial cases by perturbing real traffic scenarios. The challenge is to design a suitable threat model
such that the altered scenarios remain realistic. AdvSim~\cite{Wang2021AdvSim} plants adversarial agents that are optimized to jeopardize the ego vehicles by causing collisions, uncomfortable driving, etc. Although AdvSim enforces the dynamic feasibility of the synthesized trajectories, it uses black-box optimization which is slow and unreliable.
Our work is most similar to a very recent work~\cite{Zhang2022advpred}. However, as we will show empirically, \cite{Zhang2022advpred} fails to generate dynamically feasible adversarial trajectories. This is because its threat model simply uses dataset statistics (e.g. speed, acceleration, heading, etc.) as the dynamic parameters, which are too coarse to be used for generating realistic trajectories.
For example, the maximum acceleration in the NuScenes dataset is over $20m/s^2$ where the maximum acceleration for a top-tier sports car is only around $10m/s^2$. In contrast, our method leverages a carefully-designed differentiable dynamic model to estimate \emph{trajectory-wise} dynamic parameters. This allows our threat model to synthesize realistic and dynamically-feasible adversarial trajectories.  

\noindent\textbf{Adversarial Robustness.}
Deep learning models are shown to be generally vulnerable to adversarial attacks~\cite{adv:carlini2019evaluating,adv:Demontis19transfer,carlini2017towards,xiao2018generating,yang2020patchattack,xie2017adversarial,huang2019universal,huang2020universal,xiang2019generating,wen2019geometry,hamdi2020advpc,xiao2019meshadv}. There is a large body of literature on improving their adversarial robustness~\cite{advt:noack2021empirical,advt:sarkar2021adversarial,adv:madry2018towards,yang2019me,xu2017feature,bafna2018thwarting,papernot2016distillation,meng2017magnet,zhang2019towards,zhang2020robust,madry2017towards,goodfellow2014explaining,wong2020fast,shafahi2019adversarial}. In the AV context, many works examine on the adversarial robustness of the perception task~\cite{adv:xie2017adversarial}, while analyzing the adversarial robustness of trajectory forecaster ~\cite{Zhang2022advpred} is rarely explored. In this work, we focus on studying the adversarial robustness in the trajectory prediction task by considering its unique properties including motion and social interaction.

\section{Problem Formulation and Challenges}
\label{sec:problem_formulation}
In this section, we introduce the trajectory prediction task and then describe the threat model and assumptions for the attack and challenges.

\noindent\textbf{Trajectory Prediction Formulation.}\label{subsec:traj_pred_formulation}
In this work, we focus on the trajectory prediction task. The goal is to model the future trajectory distribution of $N$ agents conditioned on their history states and other environment context such as maps. More specifically, a trajectory prediction model takes a sequence of observed state for each agent at a fixed time interval $\Delta t$, and outputs the predicted future trajectory for each agent. For observed time steps $t \leq 0$, we denote states of $N$ agents at time step $t$ as $\textbf{X}^t = \left( x_1^t,\dotsc, x_i^t, \dotsc, x_N^t \right)$, where $x_i^t$ is the state of agent $i$ at time step $t$, which includes the position and the context information. We denote the history of all agents over $H$ {\em observed} time steps as $\textbf{X} = \left( \textbf{X}^{-H+1}, \dotsc, \textbf{X}^0\right)$.
Similarly, we denote future trajectories of all $N$ agents over $T$ {\em future} time steps as $\textbf{Y} = \left( \textbf{Y}^1,\dotsc, \textbf{Y}^T \right)$, where $\textbf{Y}^t = \left(y_1^t,\dotsc,y_N^t\right)$ denotes the states of $N$ agents at a future time step t ($t>0$). We denote the ground truth and the predicted future trajectories as $\textbf{Y}$ and $\hat{\textbf{Y}}$, respectively. A trajectory prediction model $\mathcal{P}$ aims to minimize the difference between $\hat{\textbf{Y}} = \mathcal{P}(\textbf{X})$ and $\textbf{Y}$.
In an AV stack, trajectory prediction is executed repeatedly at a fixed time interval, usually the same as $\Delta t$. We denote $L_p$ as the number of trajectory prediction being executed in several past consecutive time frames. Therefore, the histories at time frame ($-L_p < t\leq 0$) are $\textbf{X}(t) = \left( \textbf{X}^{-H-t+1}, \dotsc, \textbf{X}^{-t}\right)$, and similarly for $\textbf{Y}$ and $\hat{\textbf{Y}}$.

\noindent\textbf{Adversarial Attack Formulation.}
\label{subsec:attack_model}
In this work, we focus on the setting where an adversary vehicle (adv agent) attacks the prediction module of an ego vehicle by driving along an adversarial trajectory $\textbf{X}_{\text{adv}}(\cdot)$. The trajectory prediction model predicts the future trajectories of both the adv agent and other agents. The attack goal is to mislead the predictions at each time step and subsequently make the AV plan execute unsafe driving behaviors. As illustrated in Figure~\ref{fig:attack_overview}, by driving along a carefully crafted adversarial (history) trajectory, the trajectory prediction model predicts wrong future trajectories for both the adv agent and the other agent. The mistakes can in term lead to severe consequences such as collisions. 
 In this work, we focus on the white-box threat model,  where the adversary has access to both model parameters, history trajectories and future trajectories of all agents, to explore what a powerful adversary can do based on the Kerckhoffs's principle~\cite{shannon1949communication} to better motivate defense methods. 

\noindent\textbf{Challenges.}
\label{subsec:challenges}
The challenges of devising effective adversarial attacks against prediction modules are two-fold: (1) \textbf{Generating realistic adversarial trajectory}. In AV systems, history trajectories are generated by upstream tracking pipelines and are usually sparsely queried due to computational constraints. On the other hand, dynamic parameters like accelerations and curvatures are high order derivatives of position and are usually estimated by numerical differentiation requiring calculating difference between positions {\em within a small-time interval}. Therefore, it is difficult to estimate correct dynamic parameters from such sparsely sampled positions in the history trajectory. Without the correct dynamic parameters, it is impossible to determine whether a trajectory is realistic or not, let alone generate new trajectories. (2) \textbf{Evaluating the implications of adversarial attacks}. Most existing evaluation metrics for trajectory prediction assume benign settings and are inadequate to demonstrate the implications for AV systems under attacks. For example, a large Average Distance Error (ADE) in prediction does not directly entail concrete consequences such as collision. Therefore, we need a new evaluation pipeline to systematically determine the consequences of adversarial attacks against prediction modules to further raise the awareness of general audiences on the risk that AV systems might face.

\begin{figure}[t]
    \centering
    \includegraphics[width=0.9\textwidth]{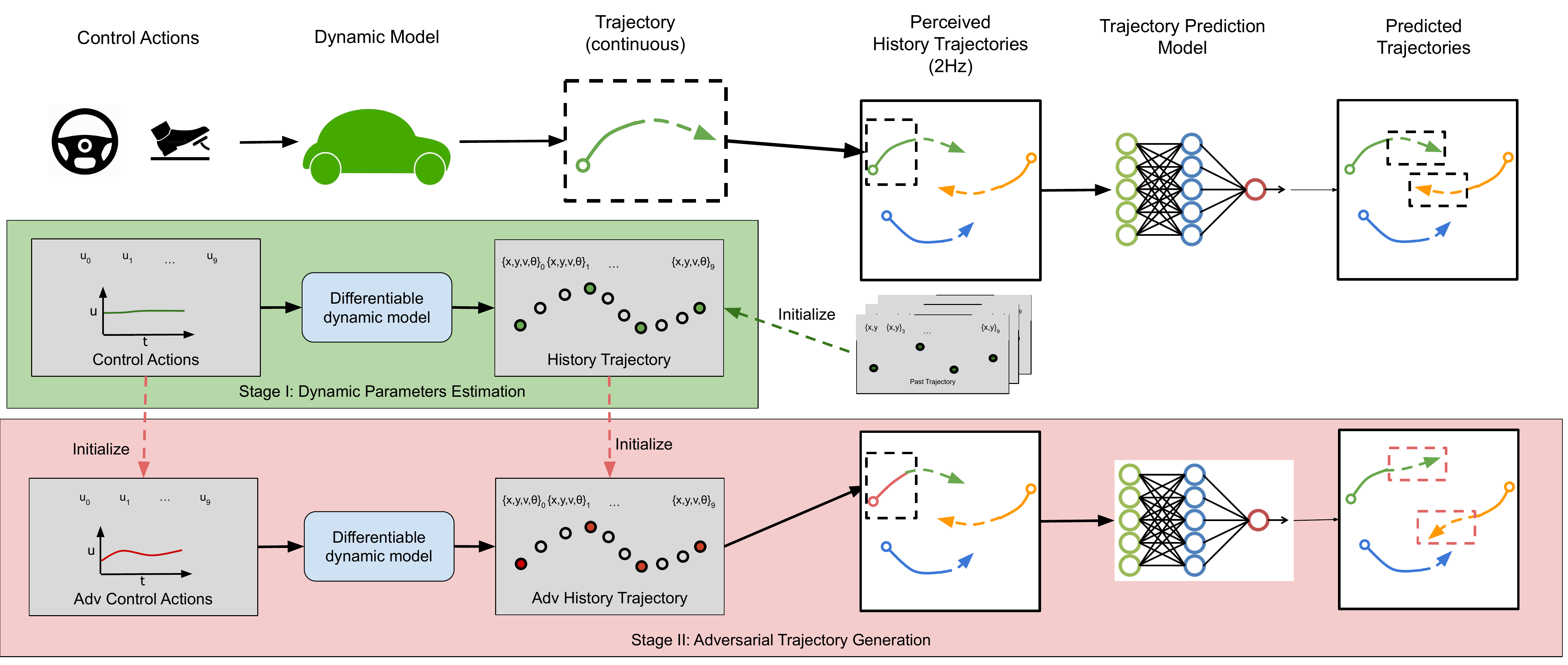}
    \caption{Adversarial Dynamic Optimization (AdvDO) methodology overview}
    \label{fig:advdo_method_overview}
\end{figure}

\section{AdvDO: Adversarial Dynamic Optimization}

To address the two challenges listed above, we propose \textbf{Adv}ersarial \textbf{D}ynamic \textbf{O}ptimization (AdvDO). As shown in Figure~\ref{fig:advdo_method_overview}, given trajectory histories, AdvDO first estimates their dynamic parameters via a differentiable dynamic model.
Then we use the estimated dynamic parameters to generate a realistic adversarial history trajectory  given a benign trajectory by solving an adversarial optimization problem. Specifically, AdvDO consists of two stages: (1) dynamic parameters estimation, and (2) adversarial trajectory generation.
In the first stage, we aim to estimate correct dynamic parameters by reconstructing a realistic dense trajectory from a sampled trajectory from the dataset. 
To reconstruct the dense trajectory, we leverage a differentiable dynamic model through optimization of control actions. When we get the estimated correct dynamic parameters of the trajectory, it could be used for the second stage. 
In the second stage, we aim to generate an adversarial trajectory that misleads future trajectory predictions given constraints. To achieve such goal, we carefully design the adversarial loss function with several regularization losses for the constraints. Then, we also extend the method to attacking consecutive predictions. 

\subsection{Dynamic Parameters Estimation}
\noindent\textbf{Differentiable dynamic model.}
A dynamic model computes the next state $s^{t+1}=\{p^{t+1},\theta^{t+1},v^{t+1}\}$ given current state $s^t = \{p^{t},\theta^{t},v^{t}\}$ and control actions $u^t=\{a^t,\kappa^t\}$. Here, $p, \theta, v, a, \kappa$ represent position, heading, speed, acceleration and curvature correspondingly. We adopt the kinematic bicycle model as the dynamic model which is commonly used
~\cite{Wang2021AdvSim}. We calculate the next state with a differential method, e.g., $v^{t+1} = v^t + a^t \cdot \Delta t$ where $\Delta t$ denotes the time difference between two time steps. Given a sequence of control actions $u=(u^0,\dotsc,u^t)$ and the initial state $s^0$, we denote the dynamic model as a differentiable function $\Phi$ such that it can calculate a sequence of future states $s = (s^0,\dotsc,s^t) = \Phi(s^0,u;\Delta t)$. Noticed that the dynamic model also provides a reverse function $\Phi^{-1}$ that calculate a sequence of dynamic parameters $\{\theta,v,a,\kappa\}=\Phi^{-1}(p;\Delta t)$ given a trajectory $p=(p^0,\dotsc,p^t)$. This discrete system can approximate the linear system in the real world when using a sufficiently small enough $\Delta t$. It can be also demonstrated that the dynamic model approximates better using a smaller $\Delta t$.

\noindent\textbf{Optimization-based trajectory reconstruction.}
\label{subsubsec:opt_reconstruction}
To accurately estimate the dynamic parameters $\{\theta,v,a,\kappa\}$ given a trajectory $p$, a small time difference $\Delta t$ or a large sampling rates $f=1/\Delta t$ is required. However, the sampling rate of the trajectory in the trajectory prediction task is decided by the AV stack, and is often small (e.g. 2Hz for nuScenes~\cite{caesar2020nuscenes}) limited by the computation performance of the hardware. 
Therefore, directly estimating  the dynamic parameters from the sampled trajectory  is not accurate, making it difficult to determine whether the adversarial history $\textbf{X}_{adv}$ generated by perturbing the history trajectory provided by the AV system is realistic or not.
To resolve this challenge, we propose to reconstruct a densely  trajectory first and then estimate a more accurate dynamic parameter from the reconstructed dense trajectory. To reconstruct a densely sampled history trajectory $\textbf{D}_i=\left( \textbf{D}^{-H \cdot f+1}_i, \dotsc, \textbf{D}^0_i\right)$ from a given history trajectory $\textbf{X}_i$ with additional sampling rates $f$, we need to find a realistic trajectory $\textbf{D}_i$ that passes through positions in $\textbf{X}_i$. We try to find it through solving an optimization problem. In order to efficiently find a realistic trajectory, we wish to optimize over the control actions in stead of the positions in $\textbf{D}_i$. To start with, we initialize $\textbf{D}_i$ with a simple linear interpolation of $\textbf{X}_i$, i.e. $\textbf{D}^{-t \cdot f+j}_i = (1-j/f) \cdot \textbf{X}^{-t} + j/f \cdot \textbf{X}^{-t+1}$. 
We then calculate the dynamic parameters for all steps $\{\theta,v,a,\kappa\}=\Phi^{-1}(\textbf{D}_i;\Delta t)$. Now, we can represent the reconstructed densely sampled trajectory $\textbf{D}_i$ with $\Phi(s^0,u;\Delta t)$, where $u=\{a,\kappa\}$. 
To further reconstruct a realistic trajectory, we optimize over the control actions $u$ with a carefully designed reconstruction loss function $\mathcal{L}_{\text{recon}}$.
The reconstruction loss function consists of two terms. We first include a MSE (Mean Square Error) loss to enforce the reconstructed trajectory passing through the given history trajectory $\textbf{X}_i$. We also include $l_{\text{dyn}}$, a regularization loss based on a soft clipping function to bound the dynamic parameters in a predefined range based on vehicle dynamics~\cite{Wang2021AdvSim}. To summarize, by solving the optimization problem of:
$$ \min_{u} \mathcal{L}_{\text{recon}}(u;s^0,\Phi)= MSE(\textbf{D}_i, \textbf{X}_i) + l_{\text{dyn}}(\theta,v,a,\kappa) $$
,we reconstruct a densely sampled, dynamically feasible trajectory $\textbf{D*}_i$ passing through the given history trajectory for the adversarial agent.

\subsection{Adversarial Trajectory Generation}
\label{subsec:adv_gen}
\noindent\textbf{Attacking a single-step prediction}.
To generate realistic adversarial trajectories, we first initialize the dynamic parameters of the adversarial agent with estimation from the previous stage, noted as $\textbf{D*}_{orig}$. Similarly to the optimization in the trajectory reconstruction process, we optimize the control actions $u$ to generate the optimal adversarial trajectories. Our adversarial optimization objective consists of four terms. The detailed formulation  for each term is in the supplementary materials.
The first term $l_{obj}$ represents the attack goal. As motion and social properties are essential and unique for trajectory prediction models. Thus, our  $l_{\text{obj}}$ has accounted for them when designed. 
The second term $l_{\text{col}}$ is a commonsense objective that encourages the generated trajectories to follow some commonsense traffic rules. In this work we only consider collision avoidance~\cite{suo2021trafficsim}. The third term $l_{\text{bh}}$ is a regularization loss based on a soft clipping function, given a clipping range of $(-\epsilon,\epsilon)$. It bounds the adversarial trajectories to be close to the original history trajectory $\textbf{X}_{\text{orig}}$. We also include $l_{\text{dyn}}$ to bound the dynamic parameters. The full adversarial loss is defined as:
$$\mathcal{L}_{\text{adv}} = l_{\text{obj}}(\textbf{Y},\hat{\textbf{Y}}) + \alpha \cdot \sum_{i} l_{\text{col}}(\textbf{D}_{\text{adv}},\textbf{X}) + \beta \cdot l_{\text{bh}}(\textbf{D}_{\text{adv}}, \textbf{D*}_{\text{orig}}) + \gamma l_{\text{dyn}}(\textbf{D}_{adv})$$
,where $\alpha$ and $\beta$ are weighting factors.
We then use the projected gradient descent (PGD) method~\cite{adv:madry2018towards} to find the adversarial control actions $u_{adv}$ bounded by constraints $(u_{lb}, u_{ub})$ attained from vehicle dynamics.

\noindent\textbf{Attacking consecutive predictions.} 
To attack $L_p$ consecutive frames of predictions, we aim to generate the adversarial trajectory of length $H+L_p$ that uniformly misleads the prediction at each time frames. 
To achieve this goal, we can easily extend the formulation for attacking single-step predictions to attack a sequence of predictions, which is useful for attacking a sequential decision maker such as an AV planning module.
Concretely, to generate the adversarial trajectories for $L_p$ consecutive steps of predictions formulated in\S~\ref{subsec:traj_pred_formulation}, we aggregate the adversarial losses over these frames. The objective for attacking a length of $H+L_p$ trajectory is:

$$\sum_{t \in [-L_p,\dotsc 0]} \mathcal{L}_{\text{adv}}(\textbf{X}(t),\textbf{D}_{\text{adv}}(t),\textbf{Y}(t))$$
, where $\textbf{X}(t),\textbf{D}_{\text{adv}}(t),\textbf{Y}(t)$ are the corresponding $\textbf{X},\textbf{D}_{\text{adv}},\textbf{Y}$ at time frame t.

\section{Experiments}
Our experiments seek to answer the following questions: (1) Are the current mainstream trajectory prediction systems  robust against our attacks?;(2) Are our attacks  more realistic compared to other methods?;   (3) How do our attacks affect an AV prediction-planning system?; (4) Does features designed to model motion and/or social properties affect a model's adversarial robustness?; and (5) Could we mitigate our attack via adversarial training? 
\subsection{Experimental Setting}
\label{subsec:exp_settings}

\noindent\textbf{Models}. We evaluate two state-of-the-art trajectory prediction models: AgentFormer and Trajectron++. As explained before, we select AgentFormer and Trajectron++ for their representative features in modeling motion and social aspects in prediction. AgentFormer proposed a transformer-based social interaction model which allows an agent’s state at one time to directly affect another agent’s state at a future time. And Trajectron++ incorporates agent dynamics.
Since semantic map is an optional information for these models, we prepare two versions for each model with map and without map.

\noindent\textbf{Datasets}. We follow the settings in~\cite{yuan2021agent,salzmann2020trajectron++} and use nuScenes dataset~\cite{caesar2020nuscenes}, a large-scale motion prediction dataset focusing on urban driving settings. We select history trajectory length ($H=4$) and future trajectory length ($T=12$) following the official recommendation. We report results on all 150 validation scenes.

\noindent\textbf{Baselines}. 
We select the search-based attack proposed by Zhang et al.~\cite{Zhang2022advpred} as the baseline, named \searchplus{}. 
As we mentioned earlier in \S~\ref{sec:related_work}, the original method made two mistakes: (1) incorrect estimated bound values for dynamic parameters and (2) incorrect choices of bounded dynamic parameters for generating realistic adversarial trajectories.
We correct such mistakes by (1) using a set of real-world dynamic bound values~\cite{Wang2021AdvSim}. 
and (2) bounding the curvature variable instead of heading derivatives since curvature is linear related to steering angle.
We denote this attack method as \search{}.
For our methods, we evaluate two variations: (1) \optinit , where the initial dynamics (i.e dynamics at $(t=-H)$ time step) $\textbf{D}^{-H\cdot S + 1}_{adv}$ are fixed and (2) \optend, where the current dynamics $(t=0)$ $\textbf{D}^0_{adv}$ are fixed. While \optend is not applicable for sequential attacks, we include \optend for understanding the attack with strict bounds, since the current position often plays an important role in trajectory prediction.

\noindent\textbf{Metrics}. We evaluate the attack with four metrics in the nuscenes prediction challenges: ADE/FDE, Miss Rates (MR), Off Road Rates (ORR)~\cite{caesar2020nuscenes} and their correspondence with planning-awareness version: PI-ADE/PI-FDE, PI-MR, PI-ORR~\cite{Ivan2022planpred} where metric values are weighted by the sensitivity to AV planning. In addition, to compare which attack method generates the most realistic adversarial trajectories, we calculate the violation rates (VR) of the curvature bound, where VR is the ratio of the number of  adversarial trajectories violating dynamics constraints over the total number of generated adversarial trajectories. 

\noindent\textbf{Implementation details}. For the trajectory reconstruction, we use the Adam optimizer and set the step number of optimization to 5. For the PGD-based attack, we set the step number to 30 for both AdvDO and baselines. We empirically choose $\beta=0.1$ and $\alpha=0.3$ for best results.

\subsection{Main Results}
\label{subsec:attack_results}

\noindent\textbf{Trajectory prediction under attacks}. First, we compare the effectiveness of the attack methods on prediction performances. 
As shown in Table~\ref{tab:attack_results}, our proposed attack (\optinit) causes the highest prediction errors across all model variants and metrics. \optinit{} overperforms \optend{} by a large margin, which shows that the dynamics of the current frame play an important role in trajectory prediction systems.
Note that \searchplus{} proposed by Zhang \emph{et al.} has a significant violation rates (VR) over 10\%. It further validates our previous claim that \searchplus{} generates unrealistic trajectories. 

\begin{table}[h]
\caption{Attack evaluation results on general metrics.}
\label{tab:attack_results}
\centering
\begin{tabular}{llrrrrc}
\toprule
Model                                  & Attack                                    & \multicolumn{1}{l}{ADE} & \multicolumn{1}{l}{FDE} & \multicolumn{1}{l}{MR} & \multicolumn{1}{l}{ORR} & Violations                \\ \midrule
                                       & None                                      & 1.83                    & 3.81                    & 28.2\%                 & 4.7\%                   & 0\%                         \\
                                       & \searchplus{}                              & 2.34                    & 4.78                    & 34.3\%                 & 6.6\%                   & 10\% \\
                                       & \search{}                                    & 1.88                    & 3.89                    & 29.2\%                 & 4.8\%                   & 0\%                         \\
                                      & {\optend{}}  & {2.23}           & 4.54           & 34.5\%        & 6.3\%          & 0\%                         \\
\multirow{-5}{*}{Agentformer w/ map}   & \textbf{\optinit{}}                        & \textbf{3.39}           & \textbf{5.75}           & \textbf{44.0\%}        & \textbf{10.4\%}         & 0\%                                   \\\midrule
                                       & None                                      & 2.20                    & 4.82                    & 35.0\%                 & 7.3\%                   & 0\%                         \\
                                       & \searchplus                              & 2.66                    & 5.53                    & 40.3\%                 & 8.9\%                   & 9\% \\
                                       & \search                                    & 2.20                    & 4.94                    & 35.1\%                 & 7.4\%                   & 0\%                         \\
                                       & \optend{}  & {2.54}           & 5.54          & 39.3\%        & 8.8\%         & 0\%                         \\
\multirow{-5}{*}{Agentformer w/o map}  & \textbf{\optinit{}} & \textbf{3.81}           & \textbf{6.01}           & \textbf{49.8\%}        & \textbf{13.3\%}         & 0\%                                                      \\ \midrule
                                       & None                                      & 1.88                    & 4.10                    & 35.1\%                 & 7.9\%                   & 0\%                         \\
                                       & \searchplus                              & 2.53                    & 5.03                    & 44.4\%                 & 9.4\%                   & 12\% \\
                                       & \search                                    & 1.93                    & 4.26                    & 36.3\%                 & 8.3\%                   & 0\%                         \\
                                       & \optend{}  & 2.48           & 5.57           & 47.5\%        & 11.3\%         & 0\%                         \\
\multirow{-5}{*}{Trajectron++ w/ map}  & \textbf{\optinit{}}                         & \textbf{3.20}           & \textbf{8.56}           & \textbf{57.2\%}        & \textbf{15.9\%}         & 0\%                         \\\midrule
                                       & None                                      & 2.10                    & 5.00                    & 41.1\%                 & 9.6\%                   & 0\%                         \\
                                       & \searchplus                              & 2.76                    & 8.02                    & 50.5\%                 & 16.1\%                  & 14\% \\
                                       & \search                                    & 2.17                    & 5.25                    & 42.2\%                 & 10.0\%                  & 0\%                         \\
                                      & \optend{}  & {2.49}           & {7.54}           & {49.5\%}        & {14.2\%}         & 0\%                         \\ 
\multirow{-5}{*}{Trajectron++ w/o map}  & \textbf{\optinit{}}                         & \textbf{3.58}           & \textbf{9.36}           & \textbf{76.8\%}        & \textbf{17.8\%}         & 0\% \\\bottomrule
\end{tabular}
\end{table}

To further demonstrate the impact of the attacks on downstream pipelines like planning, here we report prediction performance using planning-aware metrics proposed by Ivanovic \emph{et al.}~\cite{Ivan2022planpred}. As described above, these metrics consider how the predictions accuracy of surrounding agents behaviors impact the ego's ability to plan its future motion. Specifically, the metrics are computed from the partial derivative of the planning cost over the predictions to estimate the sensitivity of the ego vehicle's further planning.
Furthermore, by aggregating weighted prediction metrics (e.g., ADE, FDE, MR, ORR) with such sensitivity measurement, we could report planning awareness metrics including  (PI-ADE/FDE, PI-MR, PI-ORR) quantitatively.
As shown in Table~\ref{tab:attack_results_plan}, results are consistent with the previous results.

\begin{table}[h]
\caption{Attack evaluation results on planning-aware metrics.}
\label{tab:attack_results_plan}
\centering
\begin{tabular}{llrrrrl}
\toprule
Model                                  & Attack                                    & \multicolumn{1}{l}{PI-ADE} & \multicolumn{1}{l}{PI-FDE} & \multicolumn{1}{l}{PI-MR} & \multicolumn{1}{l}{PI-ORR} & VR                \\ \midrule
                                       & None                                       & 1.38                       & 2.76                       & 20.5\%                    & 22.8\%                     &      0\%                    \\
                                       & \searchplus{}                              & 1.62                      & 3.32                       & 25.7\%                    & 25.2\%                       & 13\% \\
                                       & \search{}                                   & 1.39                       & 2.79                       & 21.4\%                    & 23.0\%                      &          0\%                 \\
                                       & \optend                                   & 1.57                       & 3.11                       & 23.7\%                    & 24.8\%                      &            0\%               \\
\multirow{-5}{*}{Agentformer w/ map}   & \textbf{\optinit} & \textbf{2.05}              & \textbf{3.81}              & \textbf{32.9\%}           & \textbf{29.0\%}             &    0\%                       \\\midrule
                                       & None                                      & 1.46                       & 3.76                       & 26.8\%                    & 30.3\%                      &           0\%                \\
                                       & \searchplus{}                              & 1.63                       & 4.12                       & 28.9\%                    & 34.2\%                      & 11\% \\
                                       & \search{}                                  & 1.49                       & 3.74                       & 27.5\%                    & 31.1\%                      &              0\%             \\
                                       & \optend          & 1.63                       & 4.11                       & 28.2\%                    & 39.3\%                      &    0\%                       \\
\multirow{-5}{*}{Agentformer w/o map}  & \textbf{\optinit} & \textbf{2.24}              & \textbf{5.91}              & \textbf{34.3\%}           & \textbf{41.3\%}             &     0\%                      \\\midrule
                                       & None                                      & 1.42                       & 2.81                       & 26.5\%                    & 25.6\%                     &                 0\%          \\
                                       & \searchplus{}                              & 1.68                       & 3.38                       & 29.2\%                    & 28.3\%                      & 14\% \\
                                       & \search{}                                    & 1.43                       & 2.83                       & 26.7\%                    & 27.7\%                        &              0\%             \\
                                       & \optend                                   & 1.65                       & 3.14                       & 27.2\%                    & 28.1\%                      &                   0\%        \\
\multirow{-5}{*}{Trajectron++ w/ map}  & \textbf{\optinit} & \textbf{2.11}              & \textbf{3.85}              & \textbf{37.8\%}           & \textbf{32.7\%}             &            0\%               \\\midrule
                                       & None                                      & 1.76                       & 3.20                       & 30.9\%                    & 44.0\%                    &                     0\%      \\
                                       & \searchplus{}                             & 2.02                       & 3.96                       & 35.0\%                    & 49.6\%                     & 19\% \\
                                       & \search{}                                   & 1.77                       & 3.25                       & 31.0\%                    & 46.8\%                      &                  0\%         \\
                                       & \optend                                   & 1.95                       & 3.55                       & 31.6\%                    & 46.3\%                       &                    0\%       \\
\multirow{-5}{*}{Trajectron++ w/o map} & \textbf{\optinit}  & \textbf{2.46}              & \textbf{4.26}              & \textbf{41.2\%}           & \textbf{53.7\%}            &                  0\%         \\ \bottomrule
\end{tabular}
\end{table}

\begin{figure}[h]
\centering
  \includegraphics[width=0.8\textwidth]{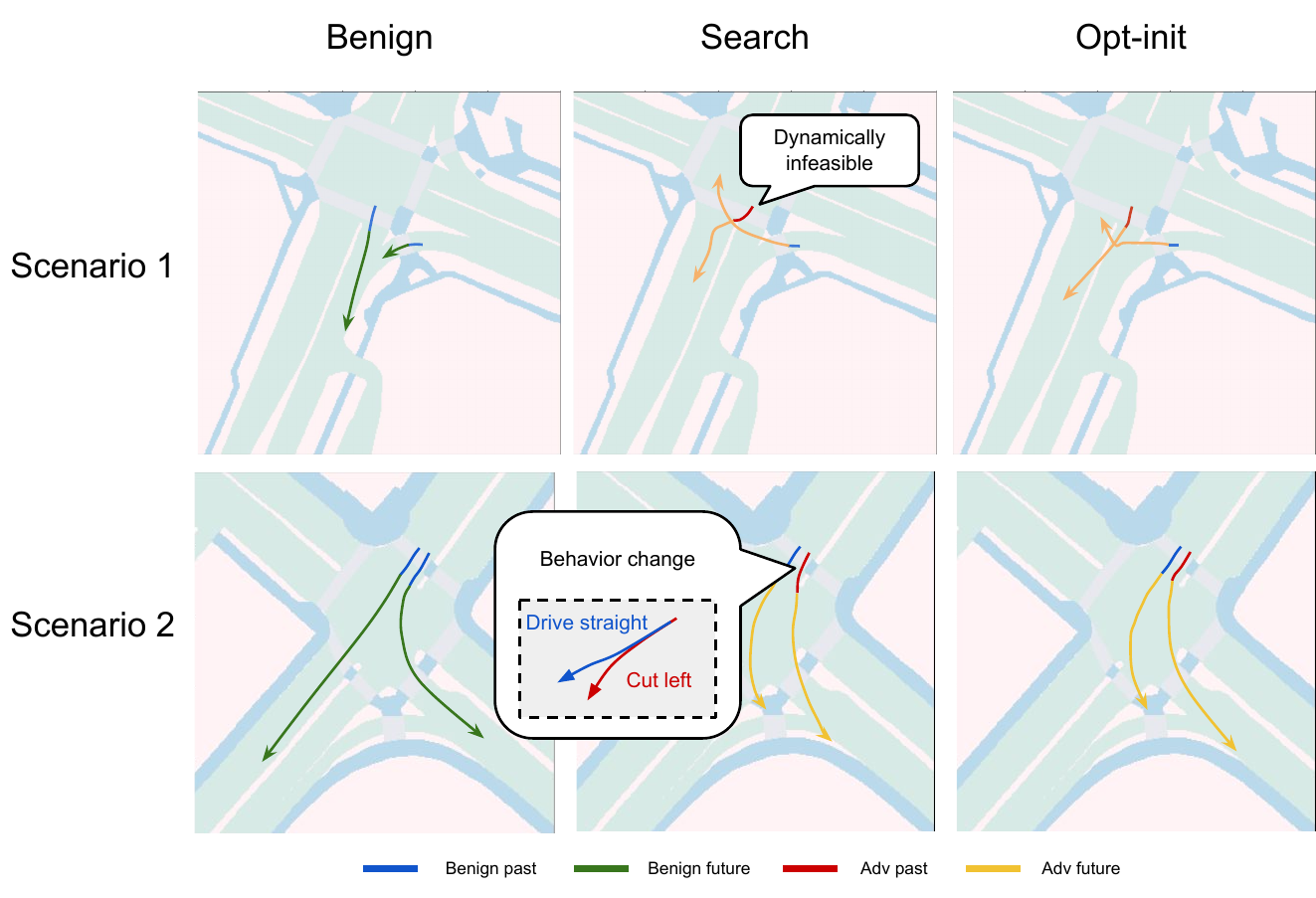}
  \caption{Qualitative comparison of generated adversarial trajectories. We demonstrate that the proposed AdvDO generates adversarial trajectories both realist and effective whereas the search-stats could either generate dynamically infeasible trajectories (sharp turn on the first row) or changing the behavior dramatically (behavior change from driving straight to swerving left on the second row).}
  \label{fig:naturalness}
\end{figure}

\begin{wraptable}{r}{0.5\textwidth}
\caption{Quantitative comparison of generated adversarial trajectories}
\label{tab:naturalness}
\centering
\begin{tabular}{cccc}
\hline\hline
Method                         & \searchplus & \optend & \optinit \\ \hline
$\Delta$Sensitivity &  2.33 &  1.12   & 1.34  \\ \hline
\end{tabular}
\end{wraptable}

\noindent\textbf{Attack fidelity analysis.}
Here, we aim to demonstrate the fidelity of the generated adversarial trajectories qualitatively and quantitatively. We show our analysis   on AgentFormer with map as a case study. In Figure~\ref{fig:naturalness}, we visualize the adversarial trajectories generated by \searchplus{} and \optend{} methods. 
We demonstrate that our method (\optend{}) can generate effective attack without changing the semantic meaning of the driving behaviors. In contrast, \searchplus{}  either generates unrealistic trajectories or changes the driving behaviors dramatically. For example, the middle row shows that the adversarial trajectory generated by \searchplus{} takes a near 90-degree sharp turn within a small distance range, which is dynamically feasible, whereas by our method (right image in the first row) generates smooth and realistic adversarial trajectories. In addition, we conduct a human study and demonstrate that only $\text{4.4}(\pm \text{2.6})\%$ of the generated adversarial trajectories are considered rule-violating. More examples of generated adversarial trajectories and details of the human study can be found in Appendix.

To further quantify the attack fidelity, we propose to use the sensitivity metric in~\cite{Ivan2022planpred} to measure the degree of behavior alteration caused by the adversarial attacks. The metric is to measure the influence of an agent's behavior over other agents' future trajectories. 
We calculate the difference of aggregated sensitivity of non-adv agents between the benign and adversarial settings. Detailed formulation is in Appendix.
We demonstrate that our proposed attacks (\textit{\optinit}, \textit{\optend}) cause smaller sensitivity changes. This corroborates our qualitative analysis that our method generates more realistic attacks at the behavior level.

\begin{table}[h]
\caption{Planning results}
\label{tab:planning_attack_results}
\centering
\begin{tabular}{l|cccc}
\toprule
\multirow{2}{*}{Planner} & \multicolumn{2}{l|}{Open-loop}        & \multicolumn{2}{l}{Closed-loop} \\ \cline{2-5} 
                         & Rule-based & \multicolumn{1}{c|}{MPC} & Rule-based         & MPC        \\ \midrule
Collisions               &  26/150  &  10/150 & 12/150 &  7/150  \\
Off road                 &    --  &   43/150   & --  & 23/150  \\ \bottomrule
\end{tabular}
\end{table}

\begin{figure}[h]
  \centering
  \begin{subfigure}[b]{0.3\textwidth}
    \includegraphics[width=\textwidth, trim={8cm 8cm 8cm 8cm},clip]{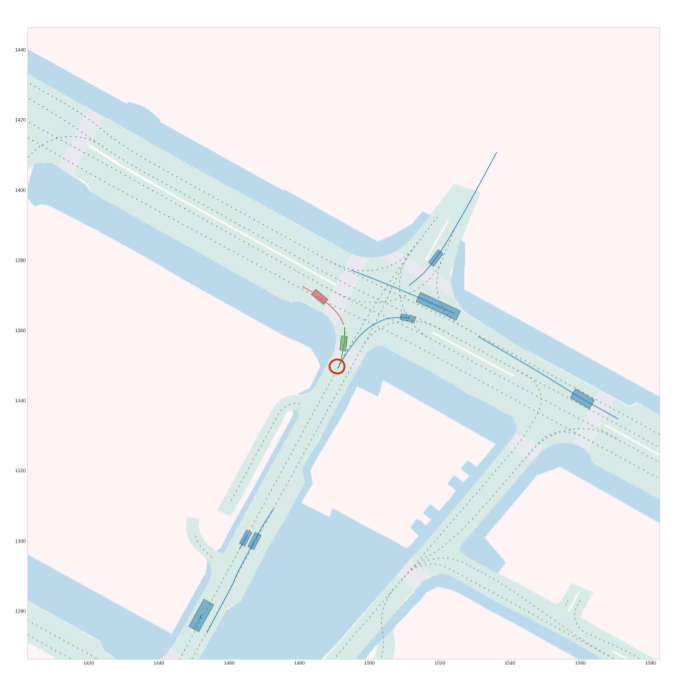}
    \caption{Side collision}
    \label{fig:planner_collision_other}
  \end{subfigure}
  \begin{subfigure}[b]{0.3\textwidth}
    \includegraphics[width=\textwidth, trim={9cm 7cm 7cm 9cm},clip]{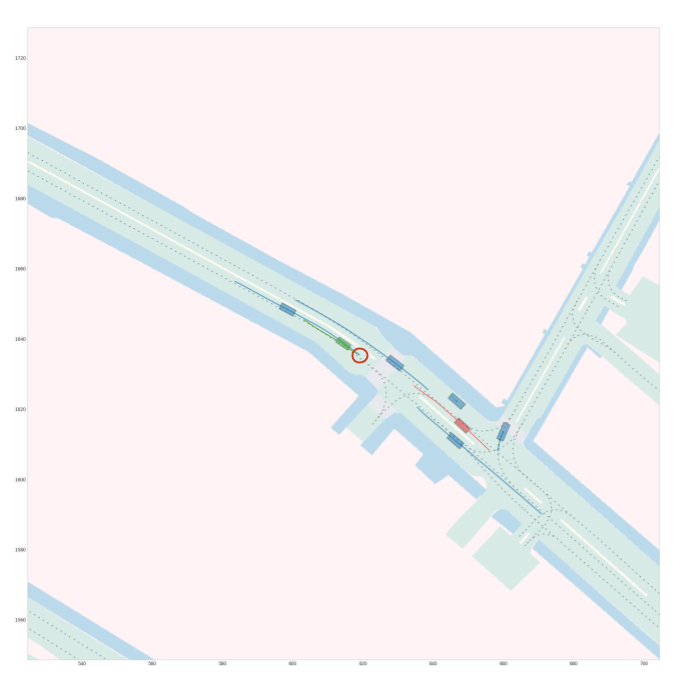}
    \caption{Rear-end collision}
    \label{fig:planner_collision_self}
  \end{subfigure}
    \begin{subfigure}[b]{0.3\textwidth}
    \includegraphics[width=\textwidth, trim={8cm 7.5cm 8cm 8.5cm},clip]{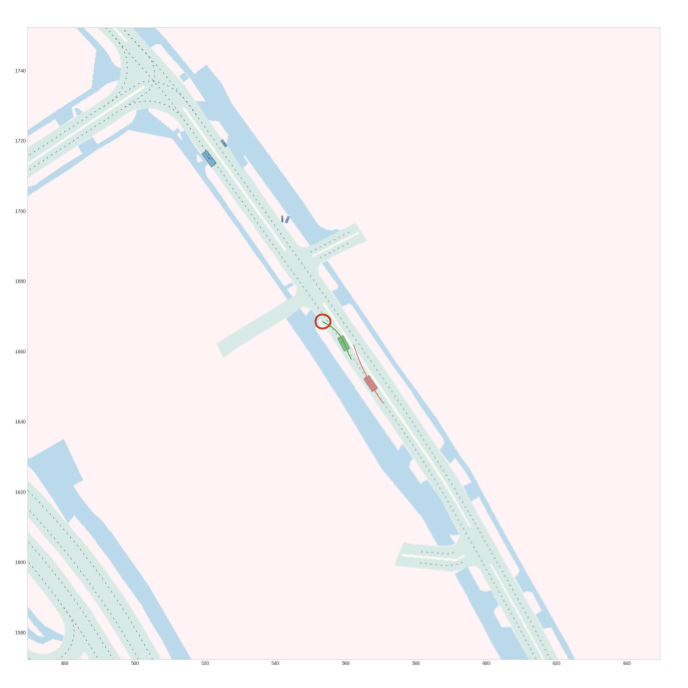}
    \caption{Driving off-road}
    \label{fig:planner_off_road}
  \end{subfigure}
  \caption{Visualized results for planner evaluation. Ego vehicle in green, adv agent in red and other agents in blue. The red cycle represents the collision or driving off-road consequence.}
  \label{fig:planner_vis}
\end{figure}

\noindent\textbf{Case studies with planners.}
\label{subsubsec:case_study}
To explicitly demonstrate the consequences of our attacks to the AV stack, we evaluate the adversarial robustness of a prediction-planning pipeline in an end-to-end manner. We select a subset of validation scenes and evaluate two planning algorithms, rule-based~\cite{rempe2021strive} and MPC-based~\cite{camacho2013model}, in in two rollout settings, open-loop and closed-loop. Detailed description for the planners can be found in Appendix. In the open-loop setting, an ego vehicle generates and follows a 6-second plan without replanning. The closed-loop setting is to replan every 0.5 seconds. We replay the other actors’ trajectories in both cases. For the closed-loop scenario, we conduct the sequential attack using $L_p = 6$. As demonstrated in Table~\ref{tab:planning_attack_results}, our attacks causes the ego to collide with other vehicles and/or leave drivable regions. We visualize a few representative cases in Figure~\ref{fig:planner_vis}. Figure~\ref{fig:planner_vis}(a) shows the attack leads to a side collision. Figure~\ref{fig:planner_vis}(b) shows the attack misleads the prediction and forces the AV to stop and leads to a rear-end collision.
Note that no attack can lead the rule-based planner to leave drivable regions because it is designed to keep the ego vehicle in the middle of the lane. At the same time, we observed that attacking the rule-based planner results in more collisions since it cannot dodge head-on collisions.

\noindent\textbf{Motion and social modeling.}
As mentioned in \S~\ref{subsec:related_work_trajectory_prediction}, trajectory prediction model aims to learn (1) the motion dynamics of each agent and (2) social interactions between agents. 
Here we conduct more in-depth attack analysis with respect to these two properties. For the motion property, we introduce a \textit{Motion} metric that measures the changes of predicted future trajectory of the adversarial agent as a result of the attack. For the social property, we hope to evaluate the influence of the attack on the predictions of non-adv agents. Thus, we use a metric named \textit{Interaction} to measure the average prediction changes among all non-adv agents. 
As shown in Table~\ref{tab:motion_interaction_ablation}, the motion property is more prone to attack than the interaction property. This is because perturbing the adv agent's history directly impacts its future, while non-adv agents are affected only through the interaction model. We observed that our attack leads to larger \textit{Motion} error for AgentFormer than for Trajectron++. A possible explanation is that AgentFormer enables direct interactions between past and future trajectories across all agents, making it more vunerable to attacks.

\begin{table}[h]
\caption{Ablation results for Motion and Interaction metrics}
\label{tab:motion_interaction_ablation}
\centering
\resizebox{\linewidth}{!}{%
\begin{tabular}{llrrrr|lrrrr}
\toprule
Model        & Scenarios   & \multicolumn{1}{l}{ADE} & \multicolumn{1}{l}{FDE} & \multicolumn{1}{l}{MR} & \multicolumn{1}{l|}{ORR} & Model & \multicolumn{1}{l}{ADE} & \multicolumn{1}{l}{FDE} & \multicolumn{1}{l}{MR} & \multicolumn{1}{l}{ORR} \\ \midrule
AgentFormer  & \textit{Motion}      & 8.12  & 12.35  & 57.3\%  & 18.6\% & Trajectron++  & 8.75  & 13.27 & 59.6\% & 16.6\%  \\
             & \textit{Interaction} & 2.03  & 4.21  & 30.3\%  & 5.1\%   & & 1.98  & 4.68  & 43.0\%  & 8.71\% \\
             
             \bottomrule
\end{tabular}}
\end{table}

\noindent\textbf{Transferability analysis.}
Here we evaluate whether the adversarial examples generated by considering one model can be transferred to attack another model. 
We report \textit{transfer rate} (more details in the appendix). 
Results are shown in Figure~\ref{fig:transfer_analysis}. Cell $(i, j)$ shows the normalized transfer rate value of adversarial examples generated against model
$j$ and evaluate on model $i$. 
We demonstrate that the generated adversarial trajectories are highly transferable (transfer rates $\geq 77\%$) when sharing the same backbone network. In addition, the generated adversarial trajectories can transfer among different backbones as well. These results show the feasibility for black-box attacks against unseen models in the real-world.

\begin{figure}[h]
  \centering
  \begin{subfigure}[b]{0.23\textwidth}
    \includegraphics[width=\textwidth,trim={1cm 0 3cm 0},clip]{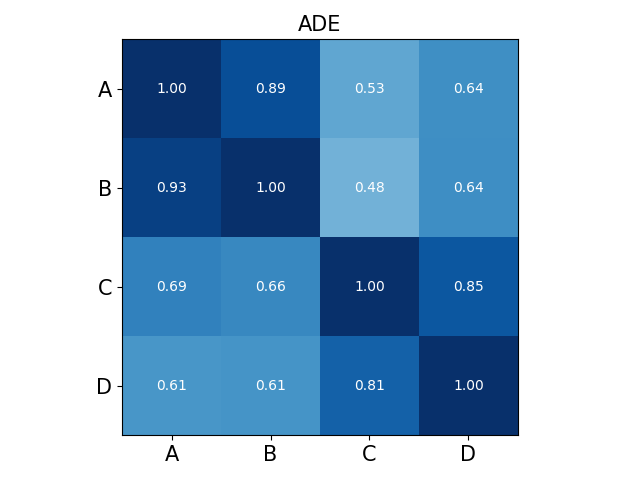}
    \caption{ADE}
    \label{fig:transfer_ADE}
  \end{subfigure}
  \begin{subfigure}[b]{0.23\textwidth}
    \includegraphics[width=\textwidth,trim={2cm 0 2cm 0},clip]{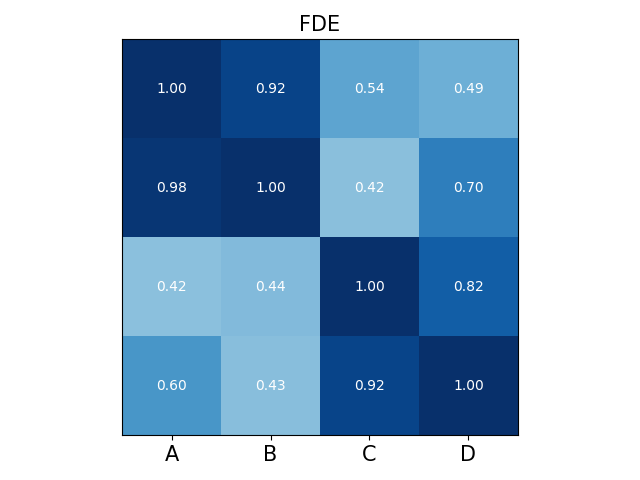}
    \caption{FDE}
    \label{fig:transfer_FDE}
  \end{subfigure}
  \begin{subfigure}[b]{0.23\textwidth}
    \includegraphics[width=\textwidth,trim={2cm 0 2cm 0},clip]{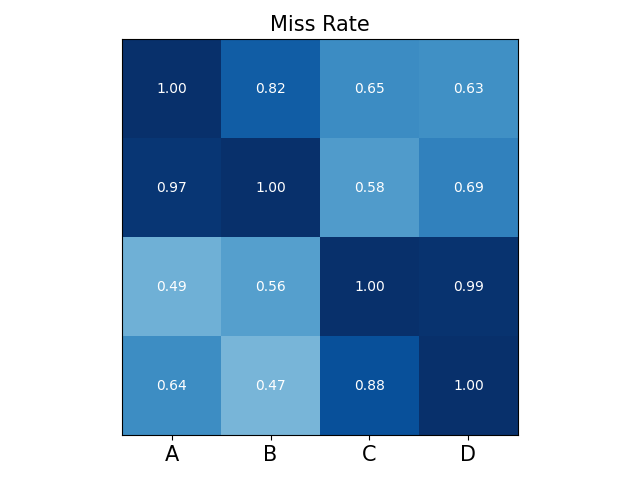}
    \caption{Miss Rate}
    \label{fig:transfer_MR}
  \end{subfigure}
  \begin{subfigure}[b]{0.23\textwidth}
    \includegraphics[width=\textwidth,trim={2cm 0 2cm 0},clip]{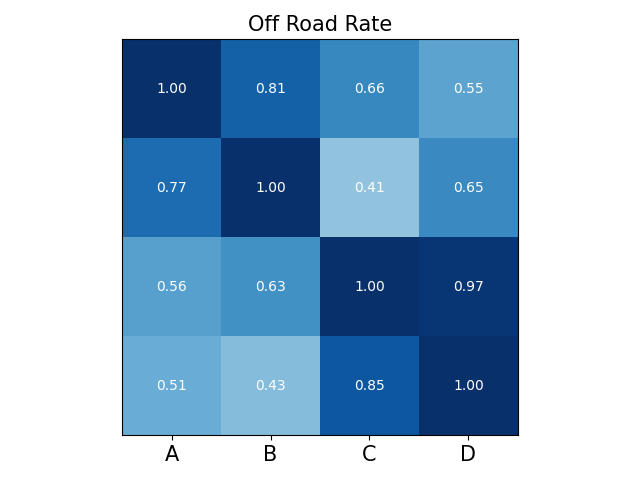}
    \caption{Off Road Rate}
    \label{fig:transfer_ORR}
  \end{subfigure}
  \caption{Transferability heatmap. A: AgentFormer w/ map; B: \& AgentFormer w/o map; C: Trajectron++ w/ map ; D: Trajectron++  \& w/o map}
  \label{fig:transfer_analysis}
\end{figure}

\noindent\textbf{Mitigation.}
To mitigate the consequences of the attacks, we use the standard mitigation method, adversarial training~\cite{adv:madry2018towards},  which has been shown as the most effective defense. 
As shown in Table C in the Appendix, we find that the adversarial trained model using the \searchplus{} attack is much worse than the adversarial trained model using our \optinit{} attack. 
This can be due to unrealistic adversarial trajectories generated by the \searchplus{}, which also emphasizes that generating realistic trajectory is essential to success of improving adversarial robustness.

\section{Conclusion}
In this paper, we study the adversarial robustness of trajectory prediction systems. We present an attack framework to generate {\em realistic} adversarial trajectories via a carefully-designed differentiable dynamic model. We have shown that prediction models are generally vulnerable and certain model designs (e.g, modeling motion and social properties simultaneously) beneficial in benign settings may make a model more vulnerable to adversarial attacks. In addition, both motion (predicted future trajectory of adversarial agent) and social (predicted future trajectory of other agents) properties could be exploited by only manipulating the adversarial agent's history trajectories. We also show that 
prediction errors influence the downstream planning and control pipeline, leading to severe consequences such as collision. We hope our study can shed light on the importance of evaluating worst-case performance under adversarial examples and raise awareness on the types of security risks that AV systems might face, so forth encourages robust trajectory prediction algorithms.


\bibliographystyle{unsrt}
\bibliography{egbib}

\begin{thebibliography}{10}

\bibitem{alahi2016social}
Alexandre Alahi, Kratarth Goel, Vignesh Ramanathan, Alexandre Robicquet,
  Li~Fei-Fei, and Silvio Savarese.
\newblock Social lstm: Human trajectory prediction in crowded spaces.
\newblock In {\em Proceedings of the IEEE conference on computer vision and
  pattern recognition}, pages 961--971, 2016.

\bibitem{ivanovic2019trajectron}
Boris Ivanovic and Marco Pavone.
\newblock The trajectron: Probabilistic multi-agent trajectory modeling with
  dynamic spatiotemporal graphs.
\newblock In {\em Proceedings of the IEEE/CVF International Conference on
  Computer Vision}, pages 2375--2384, 2019.

\bibitem{salzmann2020trajectron++}
Tim Salzmann, Boris Ivanovic, Punarjay Chakravarty, and Marco Pavone.
\newblock Trajectron++: Dynamically-feasible trajectory forecasting with
  heterogeneous data.
\newblock In {\em European Conference on Computer Vision}, pages 683--700.
  Springer, 2020.

\bibitem{yuan2021agent}
Ye~Yuan, Xinshuo Weng, Yanglan Ou, and Kris Kitani.
\newblock Agentformer: Agent-aware transformers for socio-temporal multi-agent
  forecasting.
\newblock In {\em Proceedings of the IEEE/CVF International Conference on
  Computer Vision (ICCV)}, 2021.

\bibitem{rhinehart2018r2p2}
Nicholas Rhinehart, Kris~M Kitani, and Paul Vernaza.
\newblock R2p2: A reparameterized pushforward policy for diverse, precise
  generative path forecasting.
\newblock In {\em Proceedings of the European Conference on Computer Vision
  (ECCV)}, pages 772--788, 2018.

\bibitem{rhinehart2019precog}
Nicholas Rhinehart, Rowan McAllister, Kris Kitani, and Sergey Levine.
\newblock Precog: Prediction conditioned on goals in visual multi-agent
  settings.
\newblock In {\em Proceedings of the IEEE/CVF International Conference on
  Computer Vision}, pages 2821--2830, 2019.

\bibitem{kosaraju2019social}
Vineet Kosaraju, Amir Sadeghian, Roberto Mart{\'\i}n-Mart{\'\i}n, Ian Reid,
  Hamid Rezatofighi, and Silvio Savarese.
\newblock Social-bigat: Multimodal trajectory forecasting using bicycle-gan and
  graph attention networks.
\newblock {\em Advances in Neural Information Processing Systems}, 32, 2019.

\bibitem{Ivan2022planpred}
Boris Ivanovic and Marco Pavone.
\newblock Injecting planning-awareness into prediction and detection
  evaluation.
\newblock {\em CoRR}, abs/2110.03270, 2021.

\bibitem{caesar2020nuscenes}
Holger Caesar, Varun Bankiti, Alex~H Lang, Sourabh Vora, Venice~Erin Liong,
  Qiang Xu, Anush Krishnan, Yu~Pan, Giancarlo Baldan, and Oscar Beijbom.
\newblock nuscenes: A multimodal dataset for autonomous driving.
\newblock In {\em Proceedings of the IEEE/CVF conference on computer vision and
  pattern recognition}, pages 11621--11631, 2020.

\bibitem{deo2021multimodal}
Nachiket Deo, Eric Wolff, and Oscar Beijbom.
\newblock Multimodal trajectory prediction conditioned on lane-graph
  traversals.
\newblock In {\em 5th Annual Conference on Robot Learning}, 2021.

\bibitem{suo2021trafficsim}
Simon Suo, Sebastian Regalado, Sergio Casas, and Raquel Urtasun.
\newblock Trafficsim: Learning to simulate realistic multi-agent behaviors.
\newblock In {\em Proceedings of the IEEE/CVF Conference on Computer Vision and
  Pattern Recognition}, pages 10400--10409, 2021.

\bibitem{ding2020learning}
Wenhao Ding, Baiming Chen, Minjun Xu, and Ding Zhao.
\newblock Learning to collide: An adaptive safety-critical scenarios generating
  method.
\newblock In {\em 2020 IEEE/RSJ International Conference on Intelligent Robots
  and Systems (IROS)}, pages 2243--2250. IEEE, 2020.

\bibitem{koren2019efficient}
Mark Koren and Mykel~J Kochenderfer.
\newblock Efficient autonomy validation in simulation with adaptive stress
  testing.
\newblock In {\em 2019 IEEE Intelligent Transportation Systems Conference
  (ITSC)}, pages 4178--4183. IEEE, 2019.

\bibitem{ding2021multimodal}
Wenhao Ding, Baiming Chen, Bo~Li, Kim~Ji Eun, and Ding Zhao.
\newblock Multimodal safety-critical scenarios generation for decision-making
  algorithms evaluation.
\newblock {\em IEEE Robotics and Automation Letters}, 6(2):1551--1558, 2021.

\bibitem{abeysirigoonawardena2019generating}
Yasasa Abeysirigoonawardena, Florian Shkurti, and Gregory Dudek.
\newblock Generating adversarial driving scenarios in high-fidelity simulators.
\newblock In {\em 2019 International Conference on Robotics and Automation
  (ICRA)}, pages 8271--8277. IEEE, 2019.

\bibitem{rempe2021strive}
Davis Rempe, Jonah Philion, Leonidas~J. Guibas, Sanja Fidler, and Or~Litany.
\newblock Generating useful accident-prone driving scenarios via a learned
  traffic prior.
\newblock In {\em arXiv:2112.05077}, 2021.

\bibitem{Wang2021AdvSim}
Jingkang Wang, Ava Pun, James Tu, Sivabalan Manivasagam, Abbas Sadat, Sergio
  Casas, Mengye Ren, and Raquel Urtasun.
\newblock Advsim: Generating safety-critical scenarios for self-driving
  vehicles.
\newblock {\em Conference on Computer Vision and Pattern Recognition (CVPR)},
  2021.

\bibitem{Zhang2022advpred}
Qingzhao Zhang, Shengtuo Hu, Jiachen Sun, Qi~Alfred Chen, and Z.~Morley Mao.
\newblock On adversarial robustness of trajectory prediction for autonomous
  vehicles.
\newblock {\em CoRR}, abs/2201.05057, 2022.

\bibitem{adv:carlini2019evaluating}
Nicholas Carlini, Anish Athalye, Nicolas Papernot, Wieland Brendel, Jonas
  Rauber, Dimitris Tsipras, Ian Goodfellow, Aleksander Madry, and Alexey
  Kurakin.
\newblock On evaluating adversarial robustness.
\newblock {\em arXiv preprint arXiv:1902.06705}, 2019.

\bibitem{adv:Demontis19transfer}
Ambra Demontis, Marco Melis, Maura Pintor, Matthew Jagielski, Battista Biggio,
  Alina Oprea, Cristina Nita-Rotaru, and Fabio Roli.
\newblock Why do adversarial attacks transfer? explaining transferability of
  evasion and poisoning attacks.
\newblock In {\em 28th USENIX Security Symposium (USENIX Security 19)}, pages
  321--338, Santa Clara, CA, August 2019. USENIX Association.

\bibitem{carlini2017towards}
Nicholas Carlini and David Wagner.
\newblock Towards evaluating the robustness of neural networks.
\newblock In {\em 2017 ieee symposium on security and privacy (sp)}, pages
  39--57. IEEE, 2017.

\bibitem{xiao2018generating}
Chaowei Xiao, Bo~Li, Jun-Yan Zhu, Warren He, Mingyan Liu, and Dawn Song.
\newblock Generating adversarial examples with adversarial networks.
\newblock {\em arXiv preprint arXiv:1801.02610}, 2018.

\bibitem{yang2020patchattack}
Chenglin Yang, Adam Kortylewski, Cihang Xie, Yinzhi Cao, and Alan Yuille.
\newblock Patchattack: A black-box texture-based attack with reinforcement
  learning.
\newblock In {\em European Conference on Computer Vision}, pages 681--698.
  Springer, 2020.

\bibitem{xie2017adversarial}
Cihang Xie, Jianyu Wang, Zhishuai Zhang, Yuyin Zhou, Lingxi Xie, and Alan
  Yuille.
\newblock Adversarial examples for semantic segmentation and object detection.
\newblock In {\em International Conference on Computer Vision}. IEEE, 2017.

\bibitem{huang2019universal}
Lifeng Huang, Chengying Gao, Yuyin Zhou, Cihang Xie, Alan Yuille, Changqing
  Zou, and Ning Liu.
\newblock Universal physical camouflage attacks on object detectors, 2019.

\bibitem{huang2020universal}
Lifeng Huang, Chengying Gao, Yuyin Zhou, Cihang Xie, Alan~L Yuille, Changqing
  Zou, and Ning Liu.
\newblock Universal physical camouflage attacks on object detectors.
\newblock In {\em Proceedings of the IEEE/CVF Conference on Computer Vision and
  Pattern Recognition}, pages 720--729, 2020.

\bibitem{xiang2019generating}
Chong Xiang, Charles~R Qi, and Bo~Li.
\newblock Generating 3d adversarial point clouds.
\newblock In {\em Proceedings of the IEEE/CVF Conference on Computer Vision and
  Pattern Recognition}, pages 9136--9144, 2019.

\bibitem{wen2019geometry}
Yuxin Wen, Jiehong Lin, Ke~Chen, and Kui Jia.
\newblock Geometry-aware generation of adversarial and cooperative point
  clouds.
\newblock 2019.

\bibitem{hamdi2020advpc}
Abdullah Hamdi, Sara Rojas, Ali Thabet, and Bernard Ghanem.
\newblock Advpc: Transferable adversarial perturbations on 3d point clouds.
\newblock In {\em European Conference on Computer Vision}, pages 241--257.
  Springer, 2020.

\bibitem{xiao2019meshadv}
Chaowei Xiao, Dawei Yang, Bo~Li, Jia Deng, and Mingyan Liu.
\newblock Meshadv: Adversarial meshes for visual recognition.
\newblock In {\em Proceedings of the IEEE/CVF Conference on Computer Vision and
  Pattern Recognition}, pages 6898--6907, 2019.

\bibitem{advt:noack2021empirical}
Adam Noack, Isaac Ahern, Dejing Dou, and Boyang Li.
\newblock An empirical study on the relation between network interpretability
  and adversarial robustness.
\newblock {\em SN Computer Science}, 2(1):1--13, 2021.

\bibitem{advt:sarkar2021adversarial}
Anindya Sarkar, Anirban Sarkar, Sowrya Gali, and Vineeth N~Balasubramanian.
\newblock Adversarial robustness without adversarial training: A teacher-guided
  curriculum learning approach.
\newblock {\em Advances in Neural Information Processing Systems}, 34, 2021.

\bibitem{adv:madry2018towards}
Aleksander Madry, Aleksandar Makelov, Ludwig Schmidt, Dimitris Tsipras, and
  Adrian Vladu.
\newblock Towards deep learning models resistant to adversarial attacks.
\newblock In {\em International Conference on Learning Representations}, 2018.

\bibitem{yang2019me}
Yuzhe Yang, Guo Zhang, Dina Katabi, and Zhi Xu.
\newblock Me-net: Towards effective adversarial robustness with matrix
  estimation.
\newblock {\em arXiv preprint arXiv:1905.11971}, 2019.

\bibitem{xu2017feature}
Weilin Xu, David Evans, and Yanjun Qi.
\newblock Feature squeezing: Detecting adversarial examples in deep neural
  networks.
\newblock {\em arXiv preprint arXiv:1704.01155}, 2017.

\bibitem{bafna2018thwarting}
Mitali Bafna, Jack Murtagh, and Nikhil Vyas.
\newblock Thwarting adversarial examples: An $ l\_0 $-robustsparse fourier
  transform.
\newblock {\em arXiv preprint arXiv:1812.05013}, 2018.

\bibitem{papernot2016distillation}
Nicolas Papernot, Patrick McDaniel, Xi~Wu, Somesh Jha, and Ananthram Swami.
\newblock Distillation as a defense to adversarial perturbations against deep
  neural networks.
\newblock In {\em 2016 IEEE symposium on security and privacy (SP)}, pages
  582--597. IEEE, 2016.

\bibitem{meng2017magnet}
Dongyu Meng and Hao Chen.
\newblock Magnet: a two-pronged defense against adversarial examples.
\newblock In {\em Proceedings of the 2017 ACM SIGSAC conference on computer and
  communications security}, pages 135--147, 2017.

\bibitem{zhang2019towards}
Huan Zhang, Hongge Chen, Chaowei Xiao, Sven Gowal, Robert Stanforth, Bo~Li,
  Duane Boning, and Cho-Jui Hsieh.
\newblock Towards stable and efficient training of verifiably robust neural
  networks.
\newblock {\em arXiv preprint arXiv:1906.06316}, 2019.

\bibitem{zhang2020robust}
Huan Zhang, Hongge Chen, Chaowei Xiao, Bo~Li, Duane~S Boning, and Cho-Jui
  Hsieh.
\newblock Robust deep reinforcement learning against adversarial perturbations
  on observations.
\newblock 2020.

\bibitem{madry2017towards}
Aleksander Madry, Aleksandar Makelov, Ludwig Schmidt, Dimitris Tsipras, and
  Adrian Vladu.
\newblock Towards deep learning models resistant to adversarial attacks.
\newblock {\em arXiv preprint arXiv:1706.06083}, 2017.

\bibitem{goodfellow2014explaining}
Ian~J Goodfellow, Jonathon Shlens, and Christian Szegedy.
\newblock Explaining and harnessing adversarial examples.
\newblock {\em arXiv preprint arXiv:1412.6572}, 2014.

\bibitem{wong2020fast}
Eric Wong, Leslie Rice, and J~Zico Kolter.
\newblock Fast is better than free: Revisiting adversarial training.
\newblock {\em arXiv preprint arXiv:2001.03994}, 2020.

\bibitem{shafahi2019adversarial}
Ali Shafahi, Mahyar Najibi, Amin Ghiasi, Zheng Xu, John Dickerson, Christoph
  Studer, Larry~S Davis, Gavin Taylor, and Tom Goldstein.
\newblock Adversarial training for free!
\newblock {\em arXiv preprint arXiv:1904.12843}, 2019.

\bibitem{adv:xie2017adversarial}
Cihang Xie, Jianyu Wang, Zhishuai Zhang, Yuyin Zhou, Lingxi Xie, and Alan
  Yuille.
\newblock {Adversarial Examples for Semantic Segmentation and Object
  Detection}.
\newblock In {\em IEEE International Conference on Computer Vision (ICCV)},
  2017.

\bibitem{shannon1949communication}
Claude~E Shannon.
\newblock Communication theory of secrecy systems.
\newblock {\em Bell Labs Technical Journal}, 28(4):656--715, 1949.

\bibitem{camacho2013model}
Eduardo~F Camacho and Carlos~Bordons Alba.
\newblock {\em Model predictive control}.
\newblock Springer science \& business media, 2013.

\bibitem{bicycle2017}
OA~Condrea, A~Chiru, RL~Chiriac, and S~Vlase.
\newblock Mathematical model for studying cyclist kinematics in vehicle-bicycle
  frontal collisions.
\newblock {\em {IOP} Conference Series: Materials Science and Engineering},
  252:012003, oct 2017.

\bibitem{jekel2019}
Charles~F Jekel, Gerhard Venter, Martin~P Venter, Nielen Stander, and Raphael~T
  Haftka.
\newblock {Similarity measures for identifying material parameters from
  hysteresis loops using inverse analysis}.
\newblock {\em International Journal of Material Forming}, may 2019.

\bibitem{mcnaughton2011motion}
Matthew McNaughton, Chris Urmson, John~M Dolan, and Jin-Woo Lee.
\newblock Motion planning for autonomous driving with a conformal
  spatiotemporal lattice.
\newblock In {\em 2011 IEEE International Conference on Robotics and
  Automation}, pages 4889--4895. IEEE, 2011.

\end{thebibliography}
\appendix
\renewcommand{\thetable}{\Alph{table}}
\renewcommand{\thesection}{\Alph{section}}
\renewcommand{\theequation}{S\arabic{equation}}
\renewcommand{\thefigure}{\Alph{figure}}

\section{Related works}

\begin{figure}
    \centering
    \includegraphics[width=0.6\linewidth]{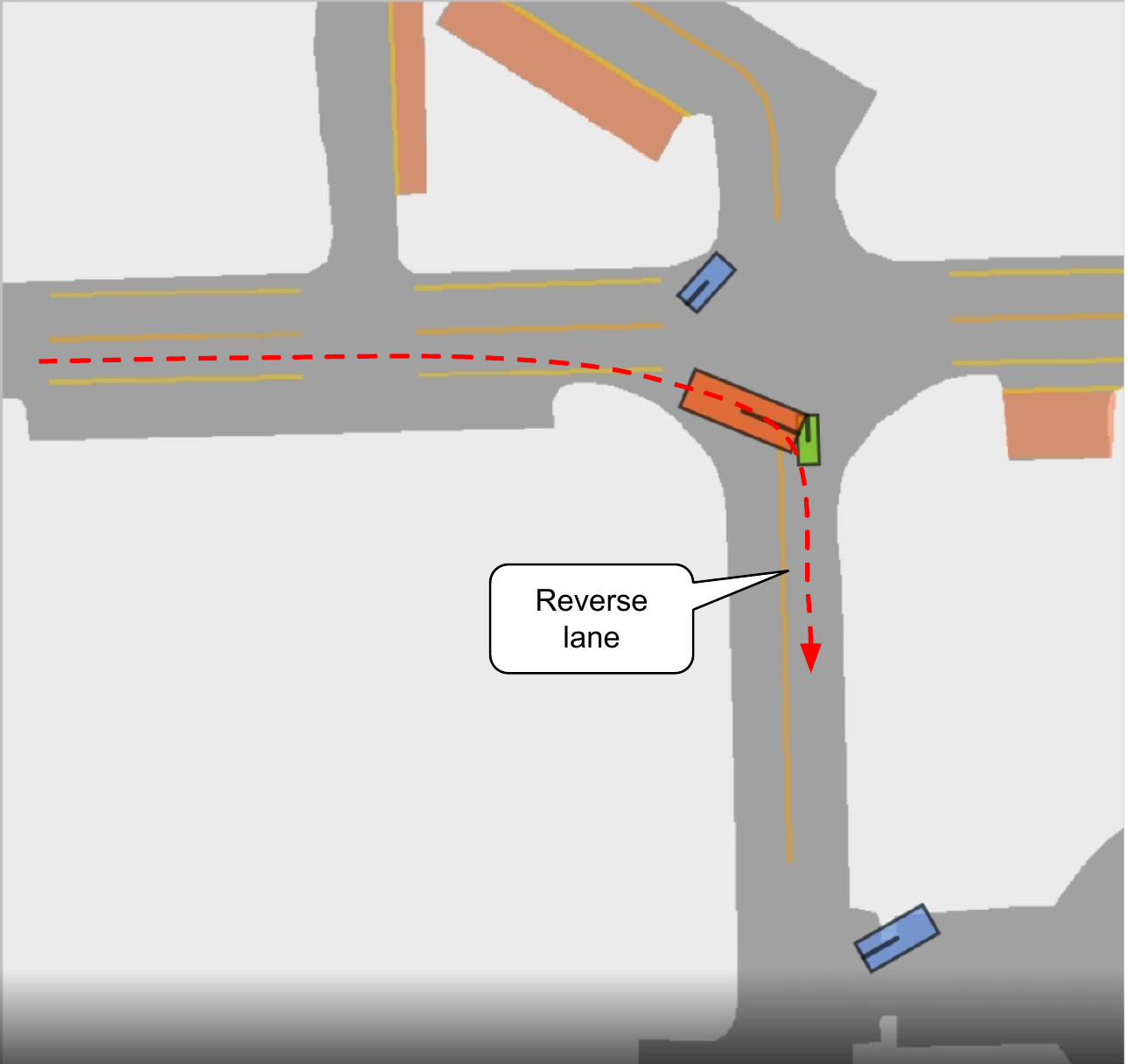}
    \caption{Adversarial agent drives in reverse lane in adversarial scenarios generatated from Strive~\cite{rempe2021strive}.}
    \label{fig:strive_demo}
\end{figure}

\textbf{Adversarial Traffic Scenarios Generation} In Strive~\cite{rempe2021strive}, adversarial scenarios generated from the traffic model is not always realistic due to the limited training data which does not cover dangerous scenarios such as collisions. In Figure~\ref{fig:strive_demo}, we demonstrated from one example generated in Strive, where the adversarial agent drives in reverse lane and violates the traffic rule, in order to collide into the AV.

\section{Method}
In this section, we describe implementation and formulation details for the proposed method.

\subsection{Differential dynamic model}
The differential dynamic model $\Phi$ is devised for deriving dynamic parameters $\{p, v, \theta\}$ from control actions $u=\{a,\kappa\}$ and deriving control actions from trajectories $p=(p_x, p_y)$. Specifically, we use a kinematic bicycle model as the dynamic model~\cite{bicycle2017}. Detailed formulation is as below:

\begin{flalign*}
    \Phi: \quad& v^{t+1} = a^t \cdot \Delta t + v^t \\
                & d\theta^{t} = v^t \cdot \kappa^t \\
                & \theta^{t+1} = d\theta^{t} \cdot \Delta t + \theta^t \\
                & p^{t+1}_x =  v^t \cdot \cos \theta^t \cdot \Delta t + p^t \\
                & p^{t+1}_y =  v^t \cdot \sin \theta^t \cdot \Delta t + p^t \\
        \Phi^{-1}: \quad& v^{t} = \Vert p^{t+1} - p^{t} \Vert/\Delta t\\
                & \theta^{t} = \arctan p_x^t/p_y^t \\
                & a^{t} = (v^{t+1} - v^t) / \Delta t \\
                & \kappa^{t} = d\theta^t / v^t
\end{flalign*}

For the physical constraints for dynamically feasibility, we follow the standard values used in~\cite{Wang2021AdvSim}.

\subsection{Reconstruction loss and adversarial loss}
Here, we describe losses for reconstruction and generating adversarial trajectory in details:
$$l_{\text{dyn}}(\theta, v, a, \kappa) = \sum_{x = \theta, v, a, \kappa} (x-x_{\text{lb}})/(x_{\text{ub}} - x_{\text{lb}}) - \text{Sigmoid}\left((x-x_{\text{lb}})/(x_{\text{ub}} - x_{\text{lb}})\right) + 0.5$$,
where $x_{\text{ub}}$, $x_{\text{lb}}$ represent the hard-coded upper bound and lower bound correspondingly for the dynamic parameter $x$.
$$l_{\text{col}}(\textbf{D}_{\text{adv}},\textbf{X}) = \frac{1}{n-1} \sum_{i\ne adv}^{n-1} \frac{1}{\Vert \textbf{D}_{\text{adv}} - \textbf{X}_i\Vert + 1}$$
, where $n$ is the number of agent in the current prediction time frame.
$$l_{\text{bh}}(\textbf{D}_{\text{adv}},\textbf{D*}_{\text{orig}},\epsilon) = \Vert\textbf{D}_{\text{adv}} - \textbf{D*}_{\text{orig}}\Vert/\epsilon - \text{Sigmoid}\left(\Vert\textbf{D}_{\text{adv}} - \textbf{D*}_{\text{orig}}\Vert/\epsilon\right) + 0.5$$
, where $\epsilon$ is the tolerance for position deviation, which we empirically set to half lane width (1 meter).
$$l_\text{obj}= \frac{1}{T} \sum_{t=1\dotsc T} \parallel \mathbf{\mathbf{Y}^t - \hat{\mathbf{Y}}^t}\parallel_2$$,
where $\hat{\mathbf{Y}}^t$ is the predicted future trajectory at time $t$ given the adversarial trajectory and $\mathbf{Y}^t$ is the corresponding ground truth. This loss aims to mislead the prediction by maximizing the difference between the predicted future trajectory and ground truth.

\section{Experiments}
In this section, we describe implementation and formulation details for the experiments.
\subsection{Attack fidelity analysis}

\begin{figure}[ht]
\begin{tabular}{ccc}
 GT \& Benign Prediction & \optend{} & \searchplus{} \\

  \includegraphics[width=0.33\linewidth,clip,trim={2cm 4cm 3cm 3cm}]{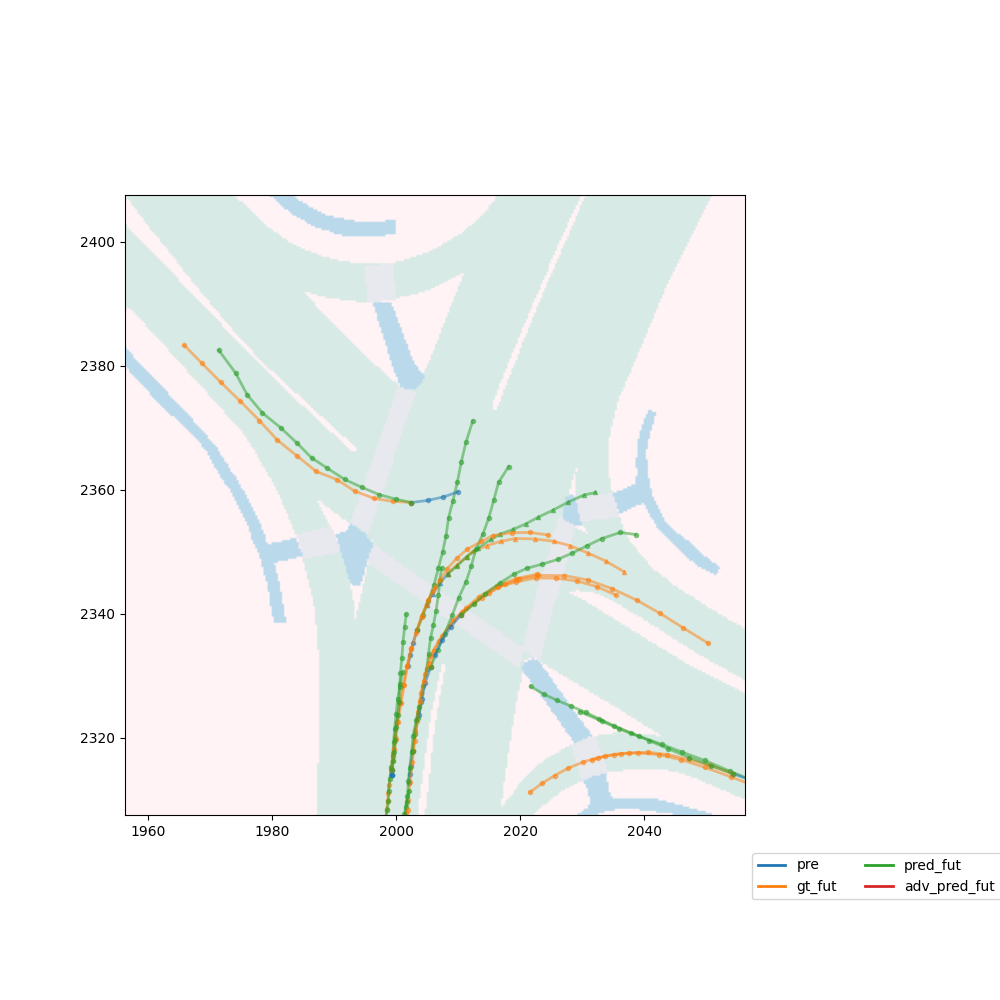} &   \includegraphics[width=0.33\linewidth,clip,trim={2cm 4cm 3cm 3cm}]{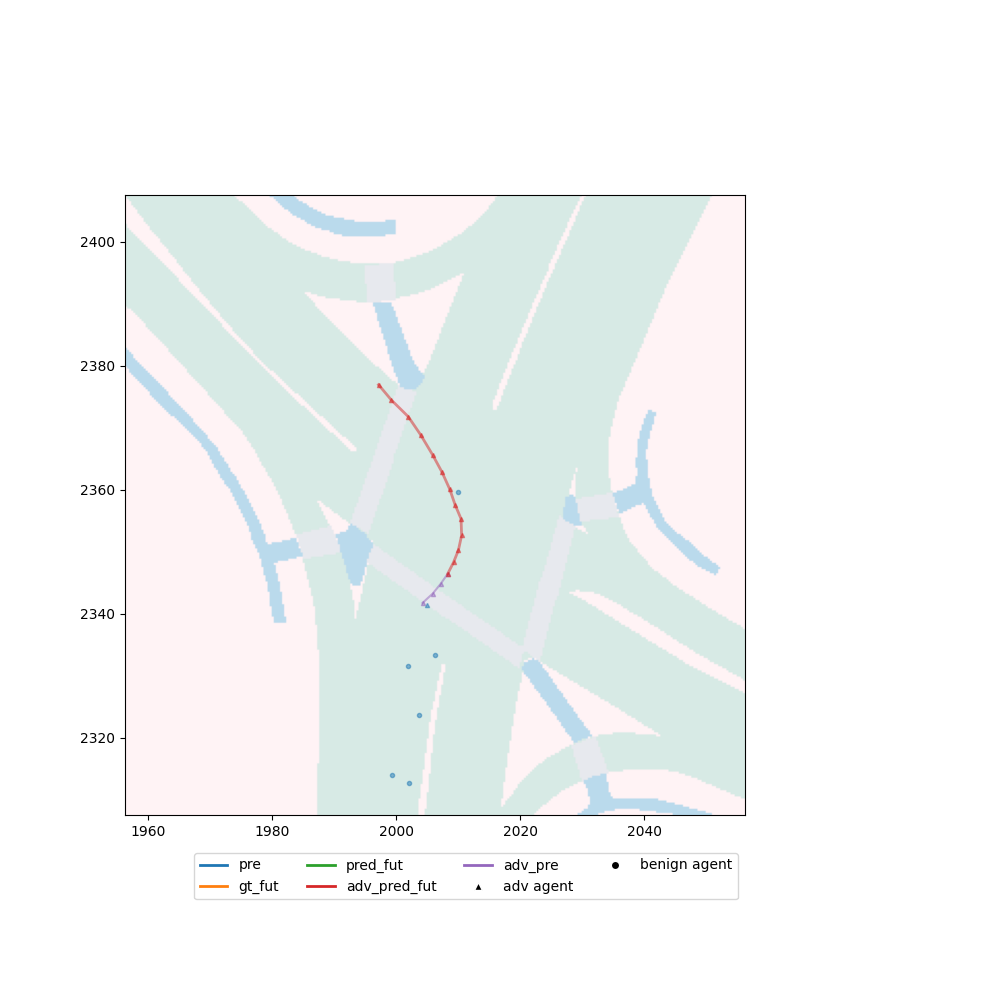} & \includegraphics[width=0.33\linewidth,clip,trim={2cm 4cm 3cm 3cm}]{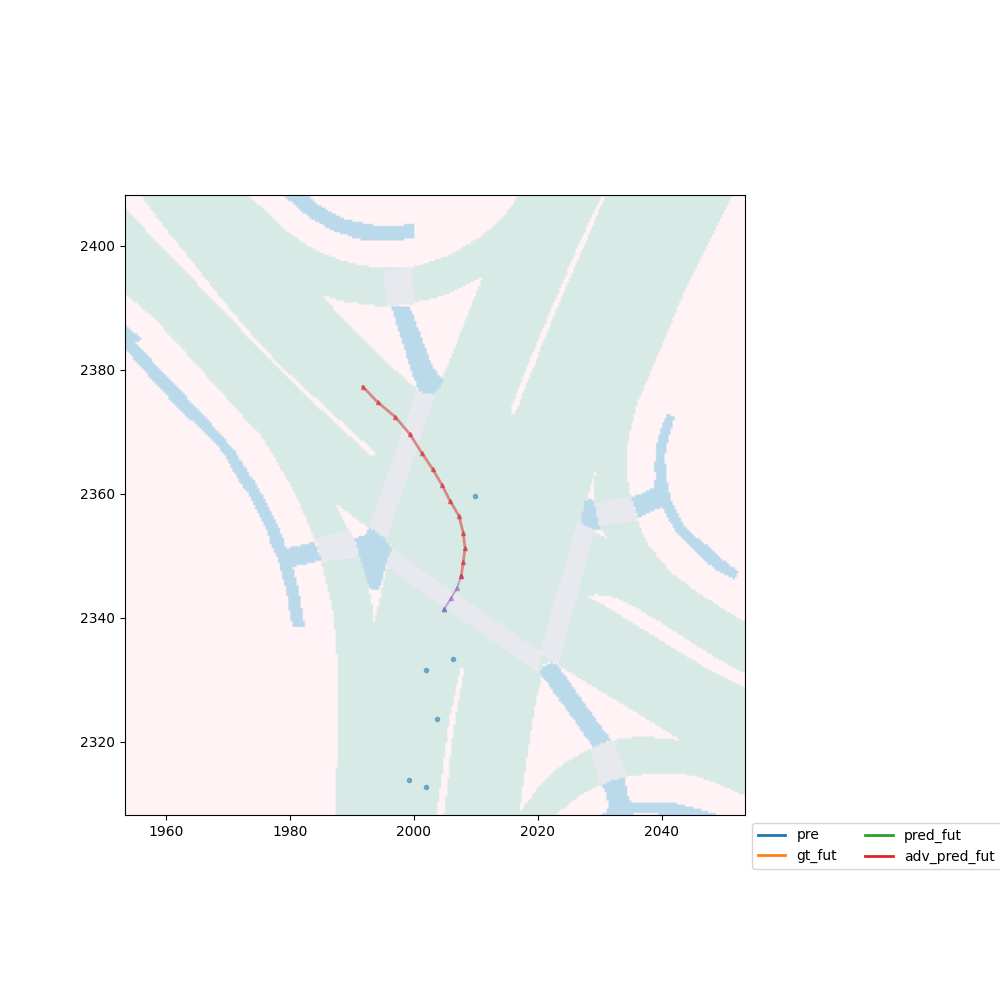} \\
  \includegraphics[width=0.33\linewidth,clip,trim={2cm 4cm 3cm 3cm}]{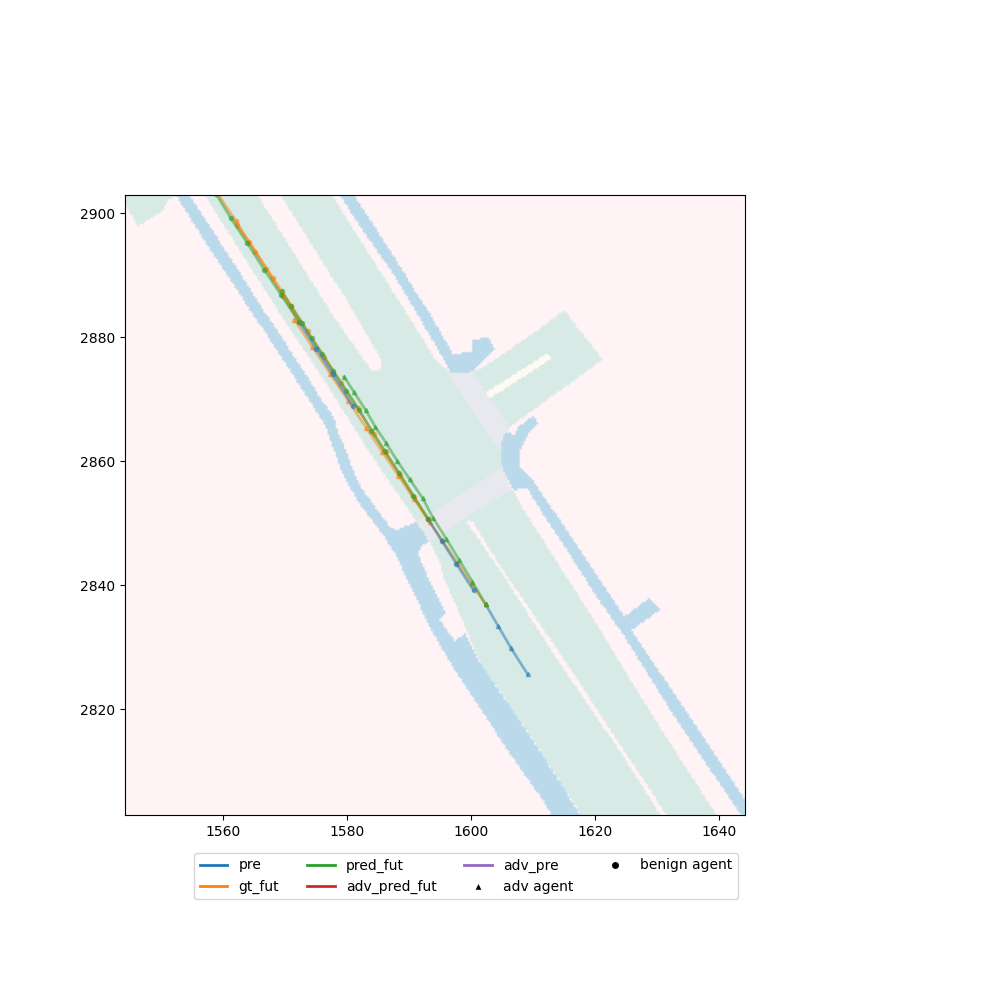} &   \includegraphics[width=0.33\linewidth,clip,trim={2cm 4cm 3cm 3cm}]{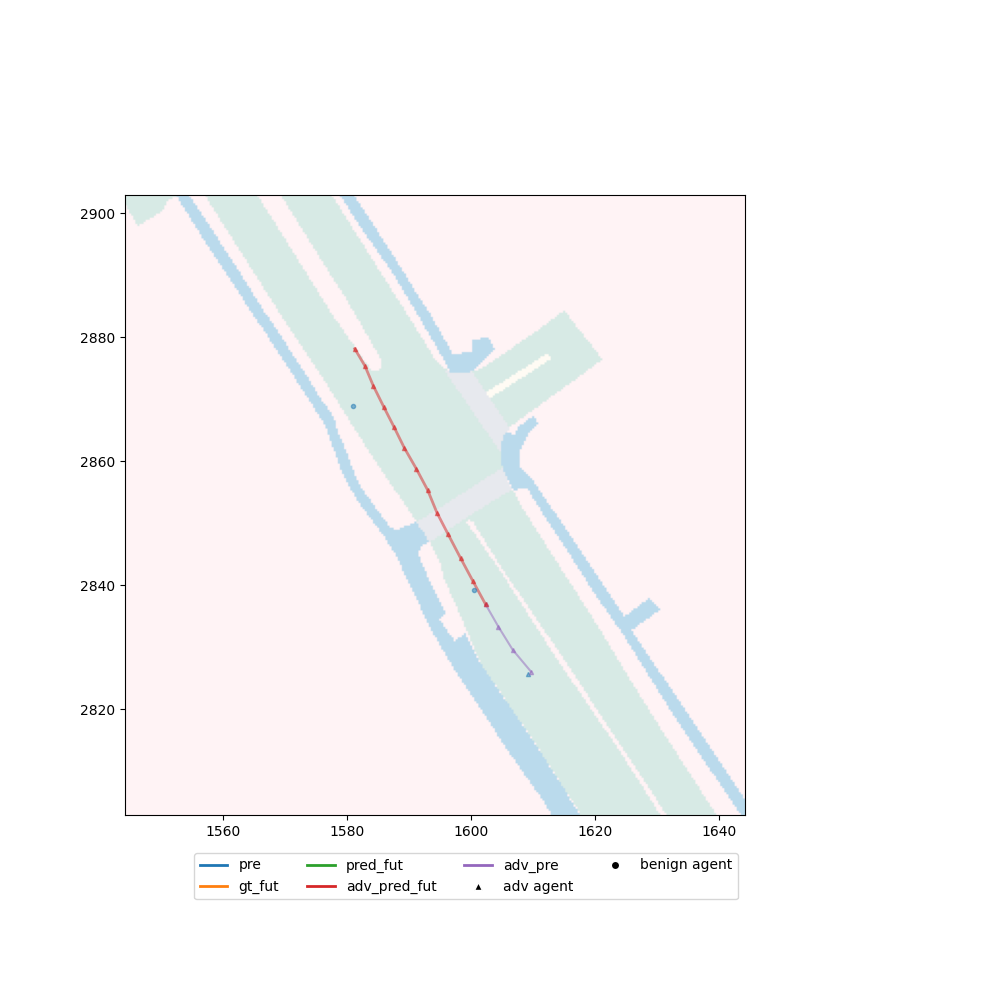} & \includegraphics[width=0.33\linewidth,clip,trim={2cm 4cm 3cm 3cm}]{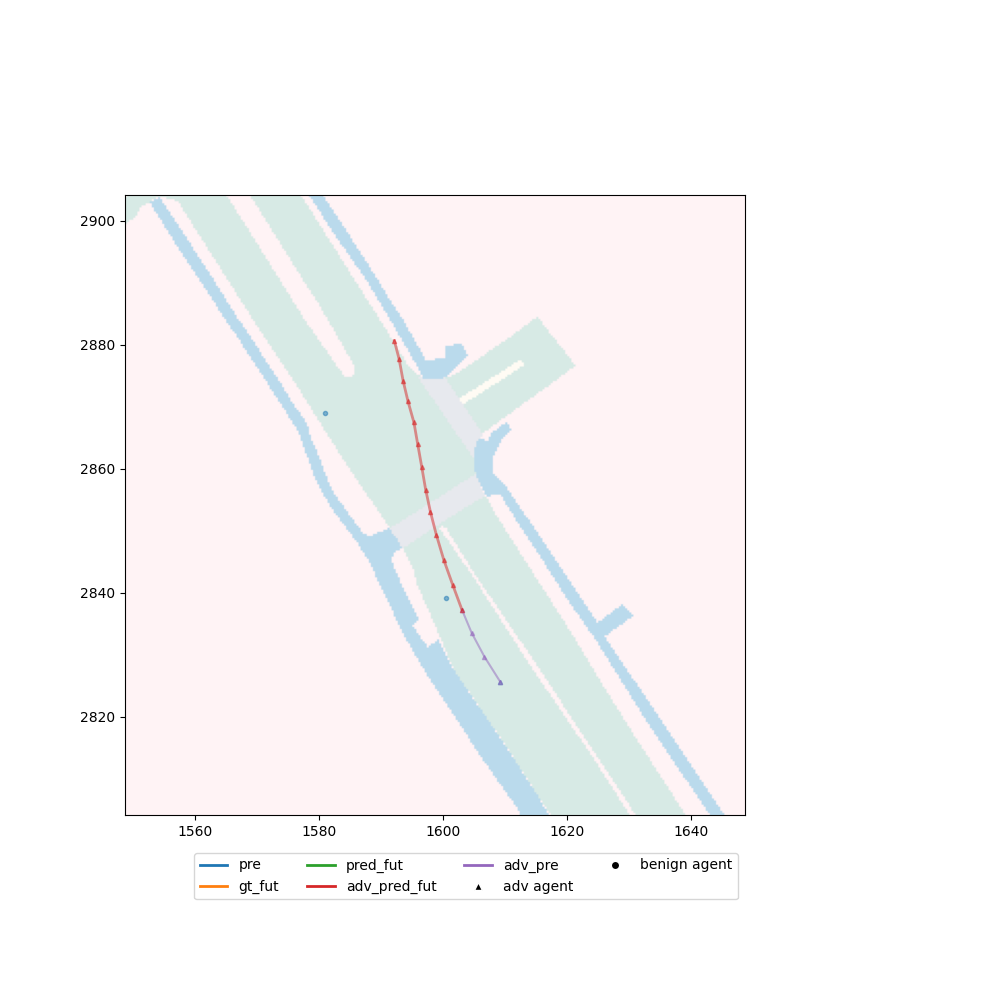} \\
  \multicolumn{3}{c}{
  \includegraphics[width=0.5\linewidth]{figs/vis_label.pdf}}
\end{tabular}
\caption{Visualization examples of generated adversarial trajectories from \optend{} and \searchplus{}. We only show the adversarial agent's trajectory in the attack scenario for clearer visualization.}
\label{fig:adv_traj_vis}
\end{figure}

In this analysis, we aim to demonstrate the generated adversarial trajectory is realistic from both perspectives of: (1) dynamically feasibility and (2) similar behavior as the original history trajectory. For the first perspective, we demonstrate the results quantitatively with the \textbf{Violation Rates} (VR) metric described below. For the second perspective, since it is a common challenge to measure the behavior change quantitatively, we propose to approximate the degree of behavior change with the \textbf{Aggregated sensitivity} metric described below. We also visually examine generated adversarial trajectories in Figure~\ref{fig:adv_traj_vis}.

\begin{figure}[ht]
\centering
\begin{tabular}{ccc}
 GT \& Benign Prediction & \optend{} & \searchplus{} \\
  \includegraphics[width=0.33\linewidth,clip,trim={2cm 4cm 3cm 3cm}]{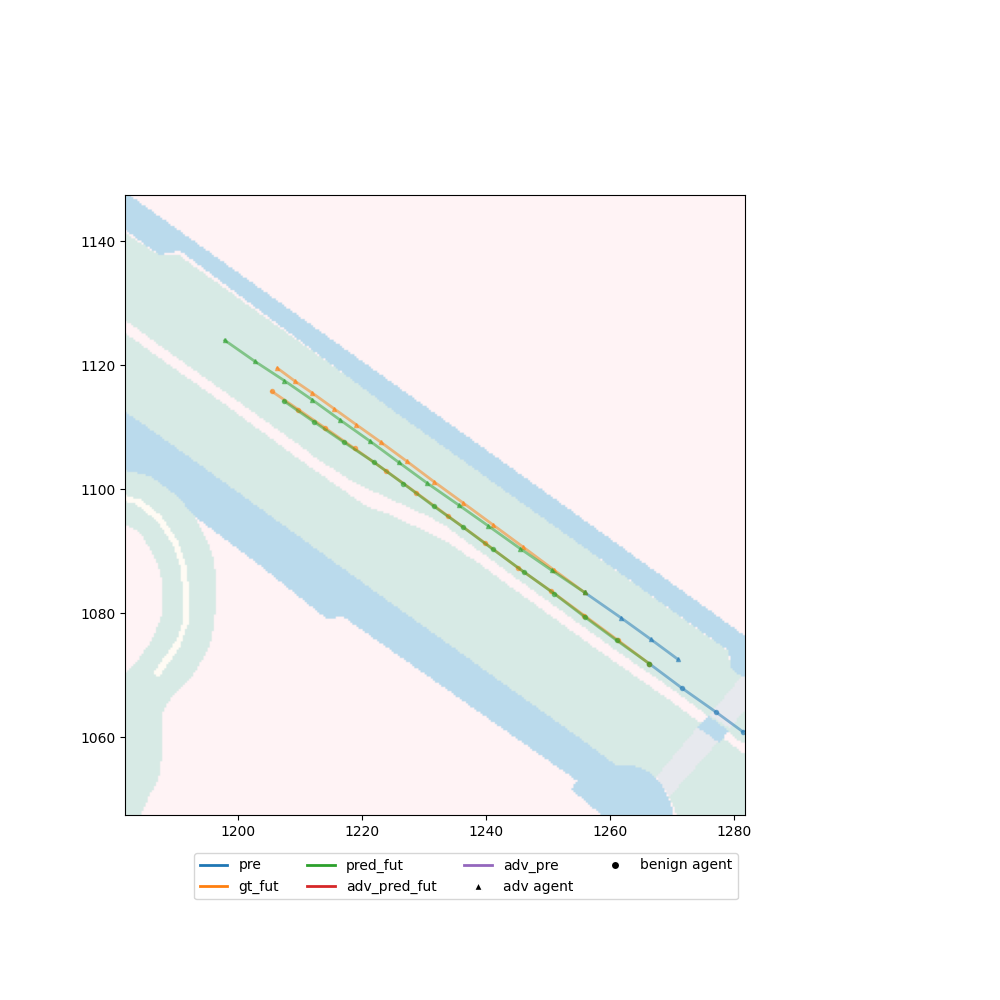} &   \includegraphics[width=0.33\linewidth,clip,trim={2cm 4cm 3cm 3cm}]{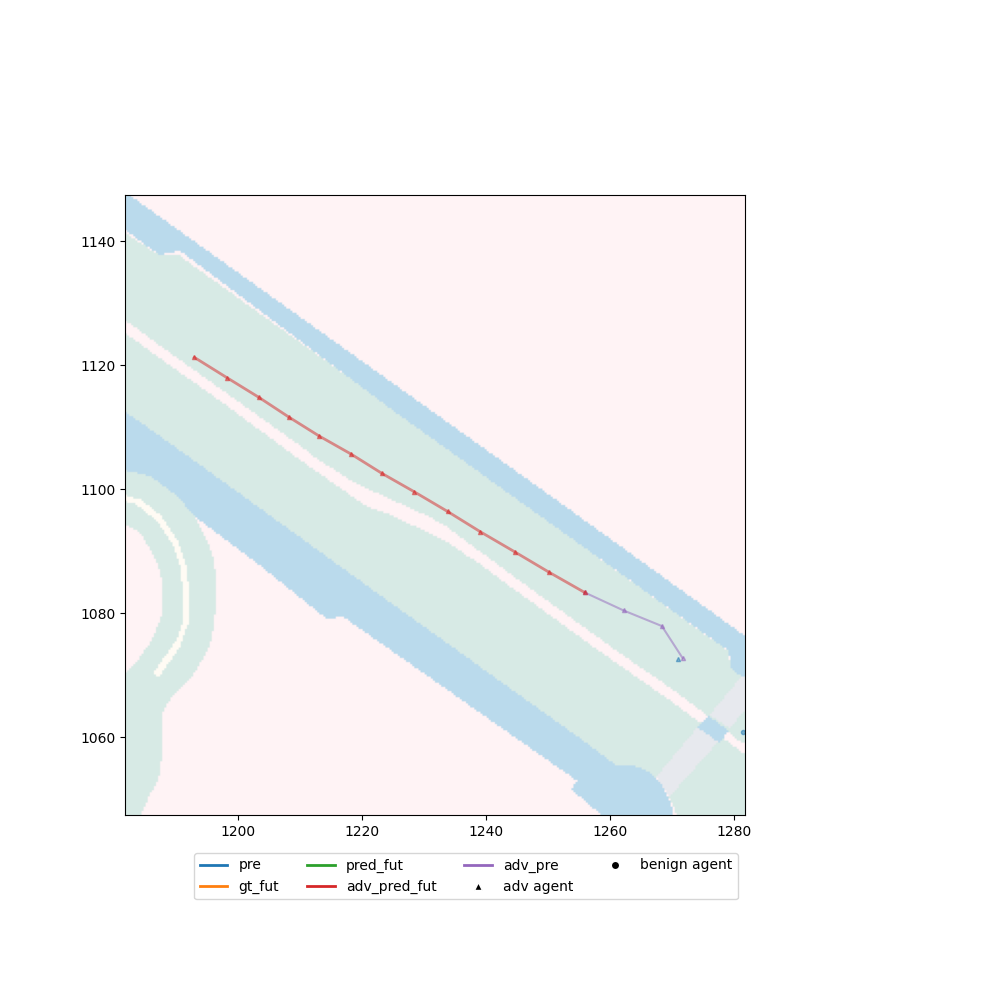} & \includegraphics[width=0.33\linewidth,clip,trim={2cm 4cm 3cm 3cm}]{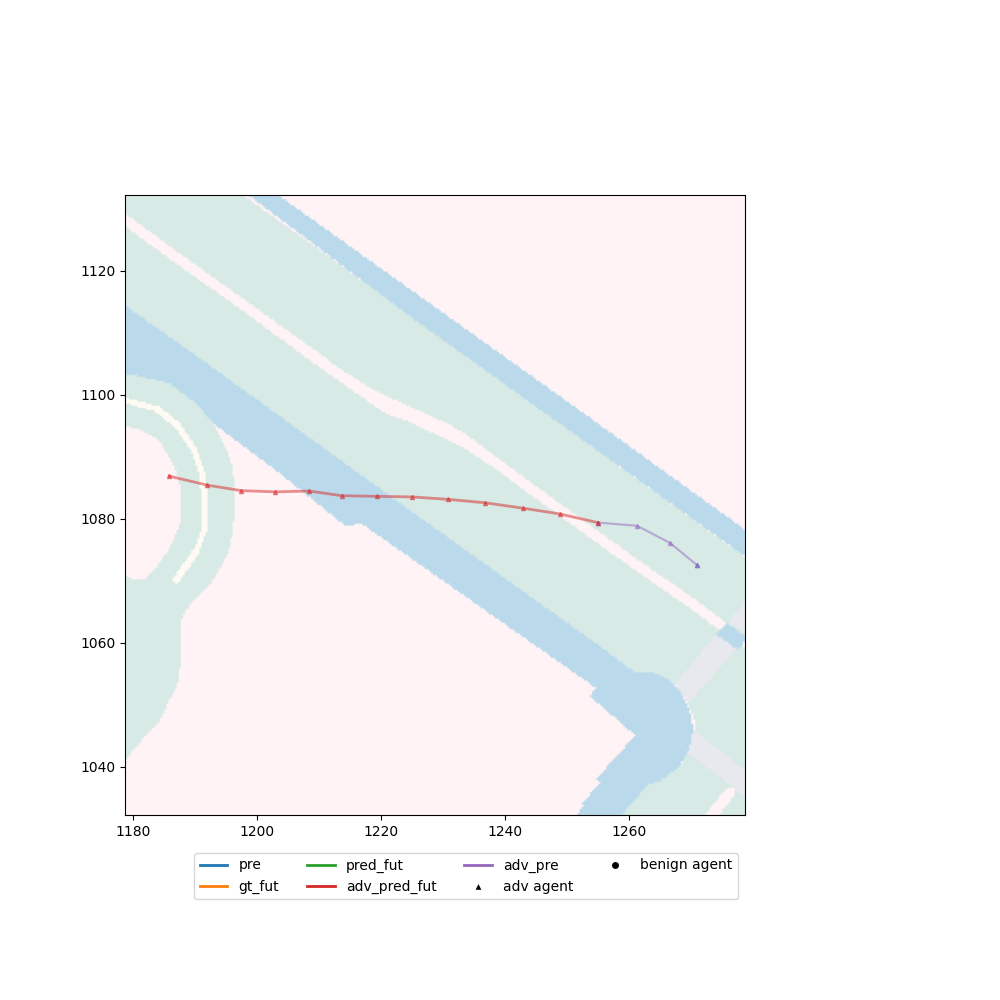} \\
  \includegraphics[width=0.33\linewidth,clip,trim={2cm 4cm 3cm 3cm}]{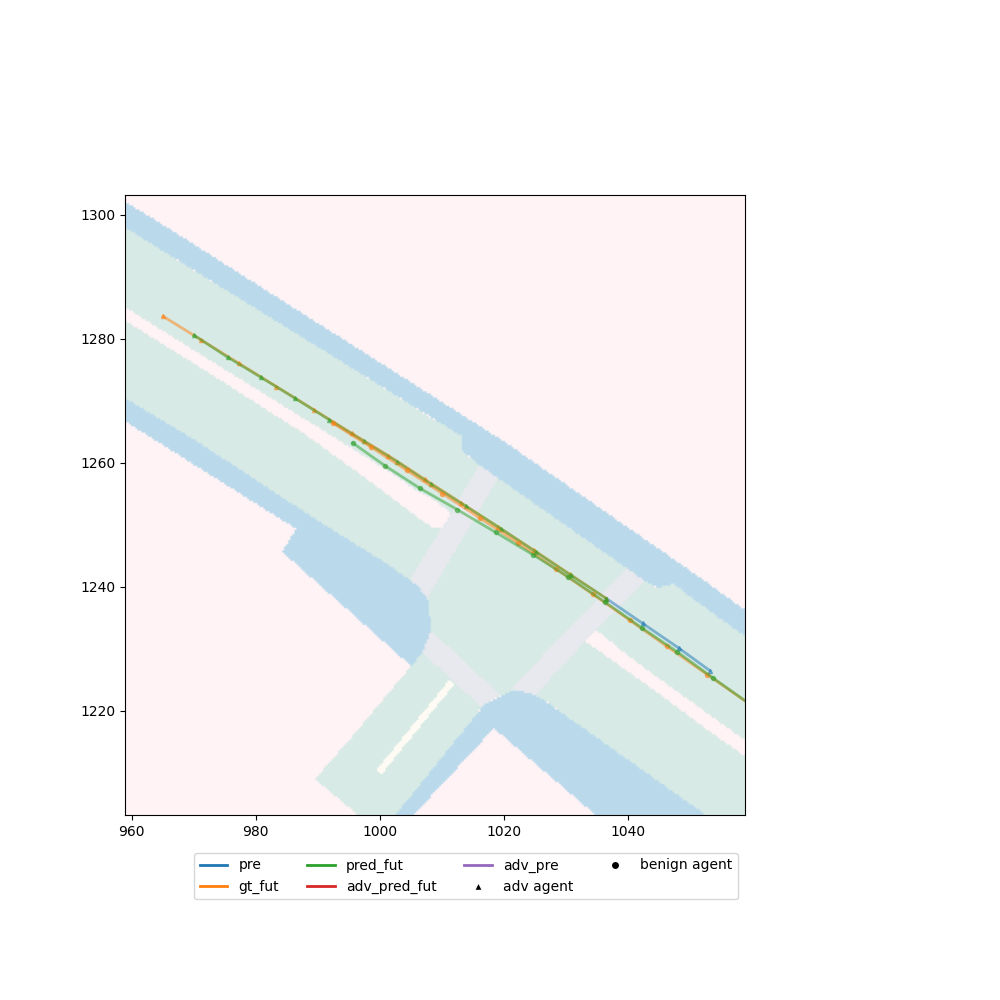} &   \includegraphics[width=0.33\linewidth,clip,trim={2cm 4cm 3cm 3cm}]{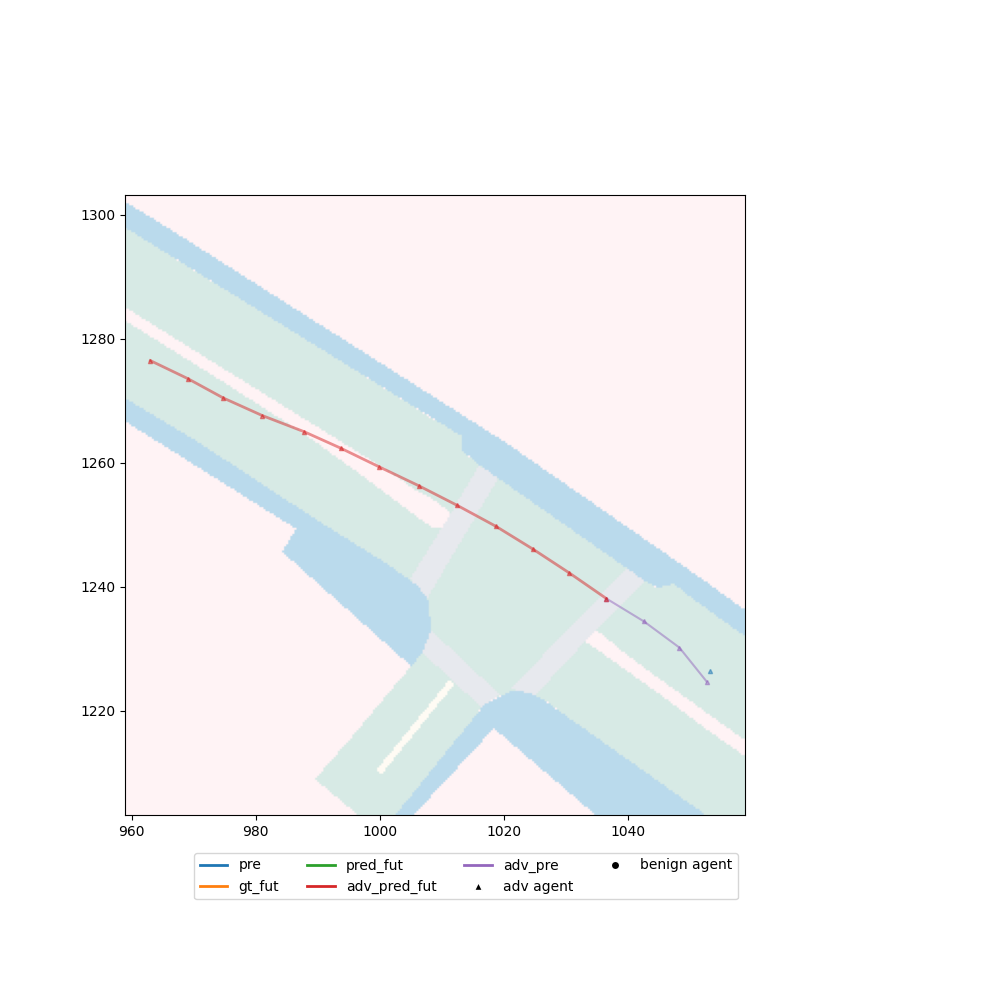} & \includegraphics[width=0.33\linewidth,clip,trim={2cm 4cm 3cm 3cm}]{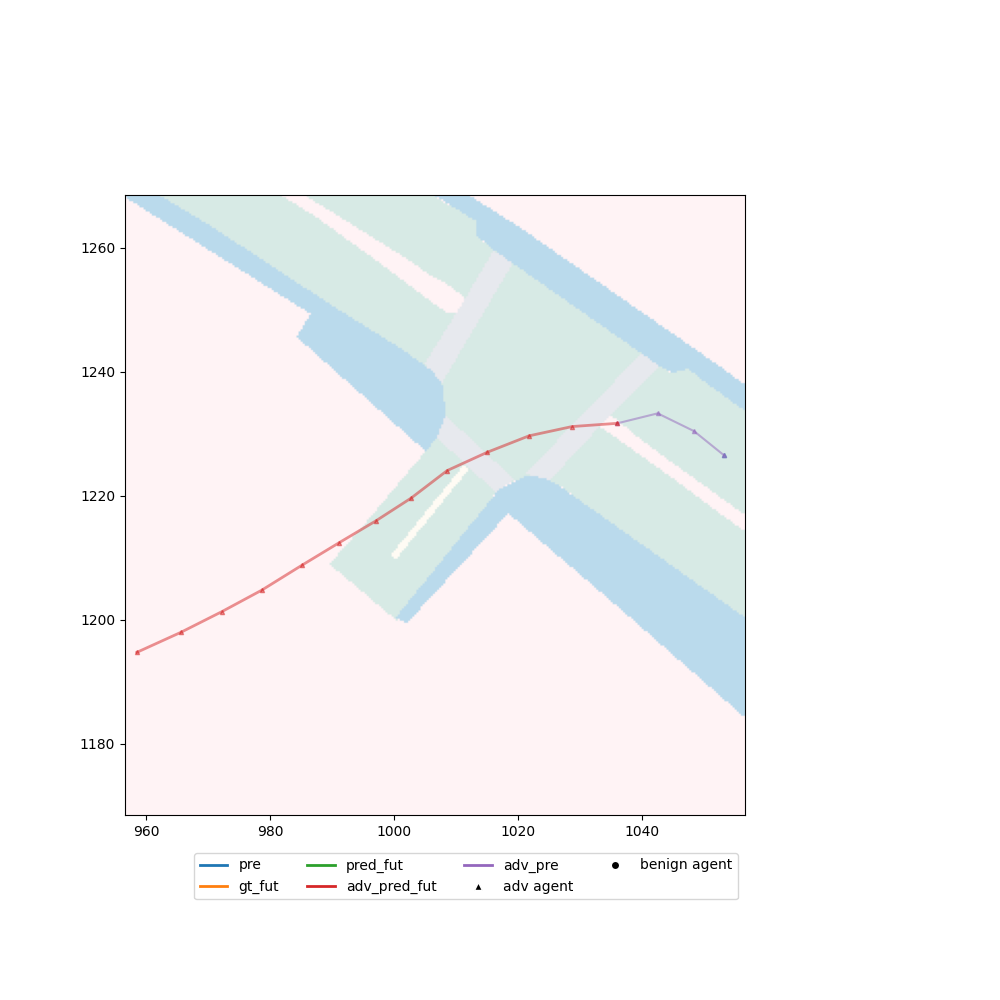} \\
  \includegraphics[width=0.33\linewidth,clip,trim={2cm 4cm 3cm 3cm}]{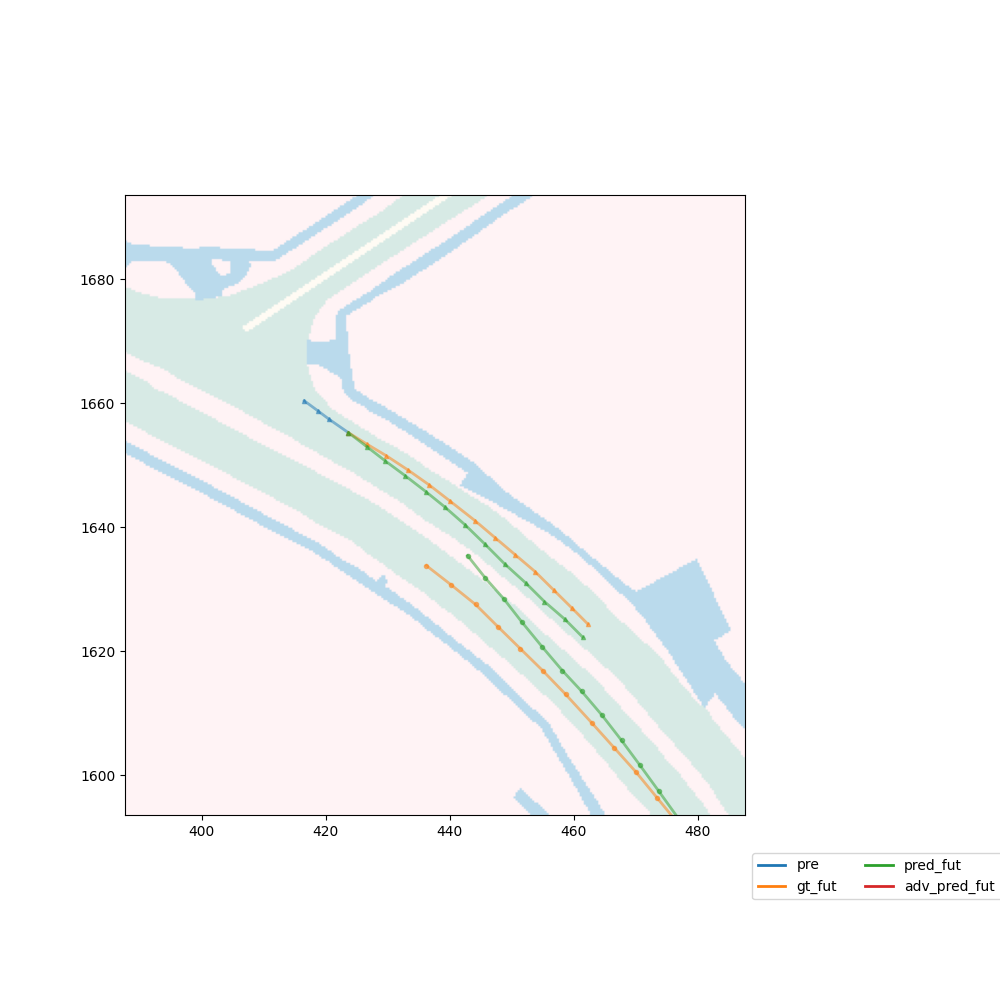} &   \includegraphics[width=0.33\linewidth,clip,trim={2cm 4cm 3cm 3cm}]{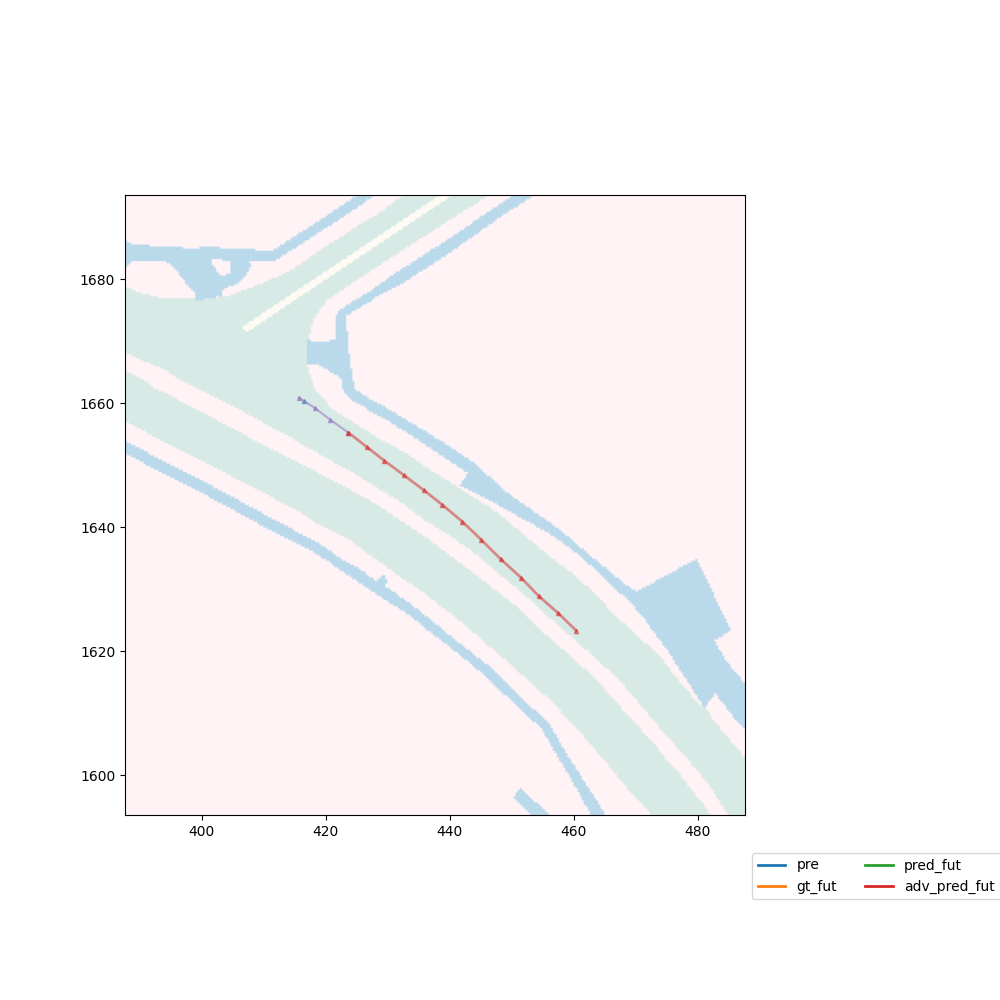} & \includegraphics[width=0.33\linewidth,clip,trim={2cm 4cm 3cm 3cm}]{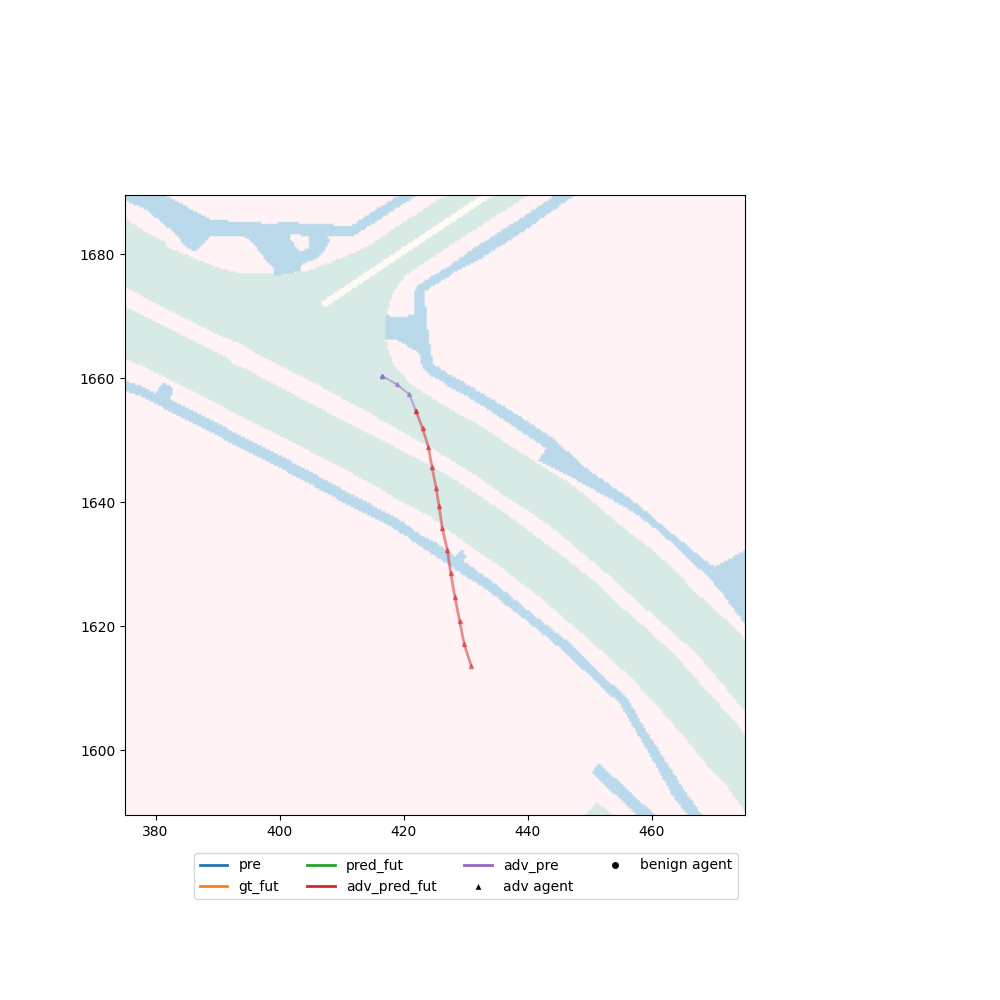} \\
  \includegraphics[width=0.33\linewidth,clip,trim={2cm 4cm 3cm 3cm}]{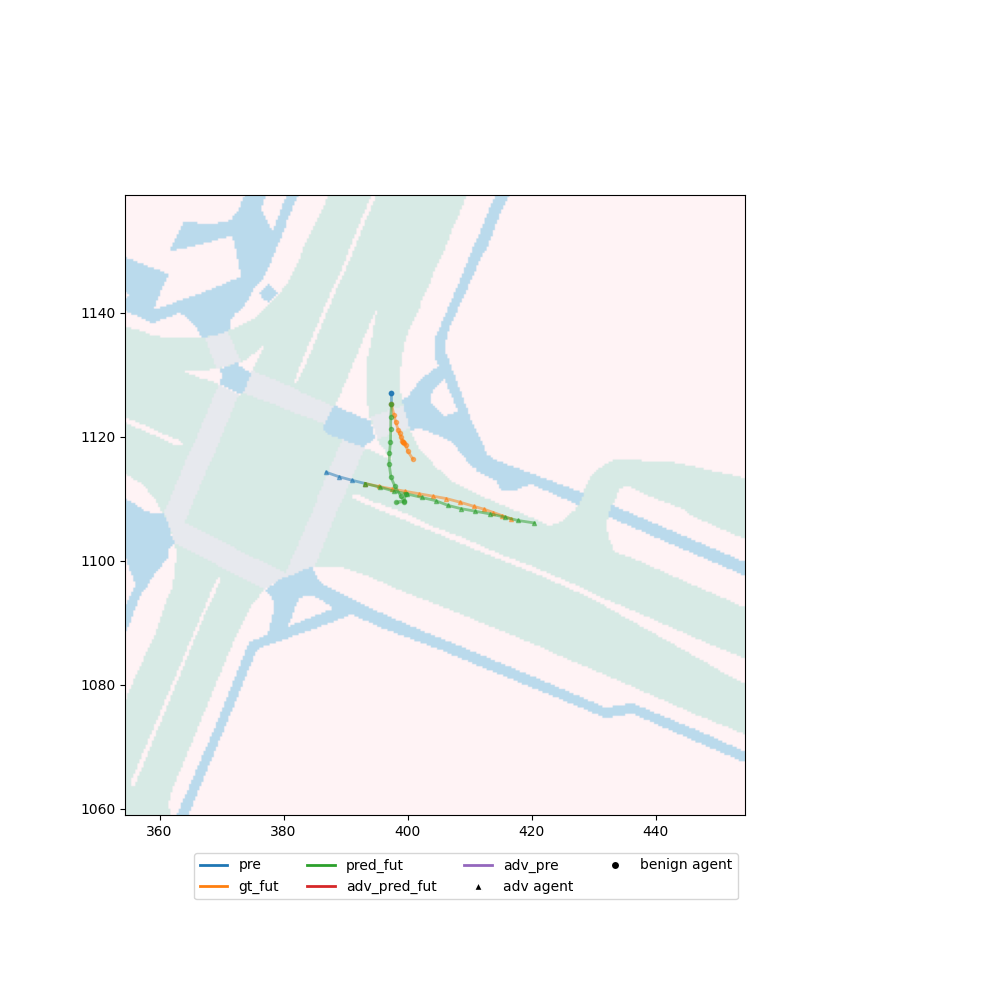} &   \includegraphics[width=0.33\linewidth,clip,trim={2cm 4cm 3cm 3cm}]{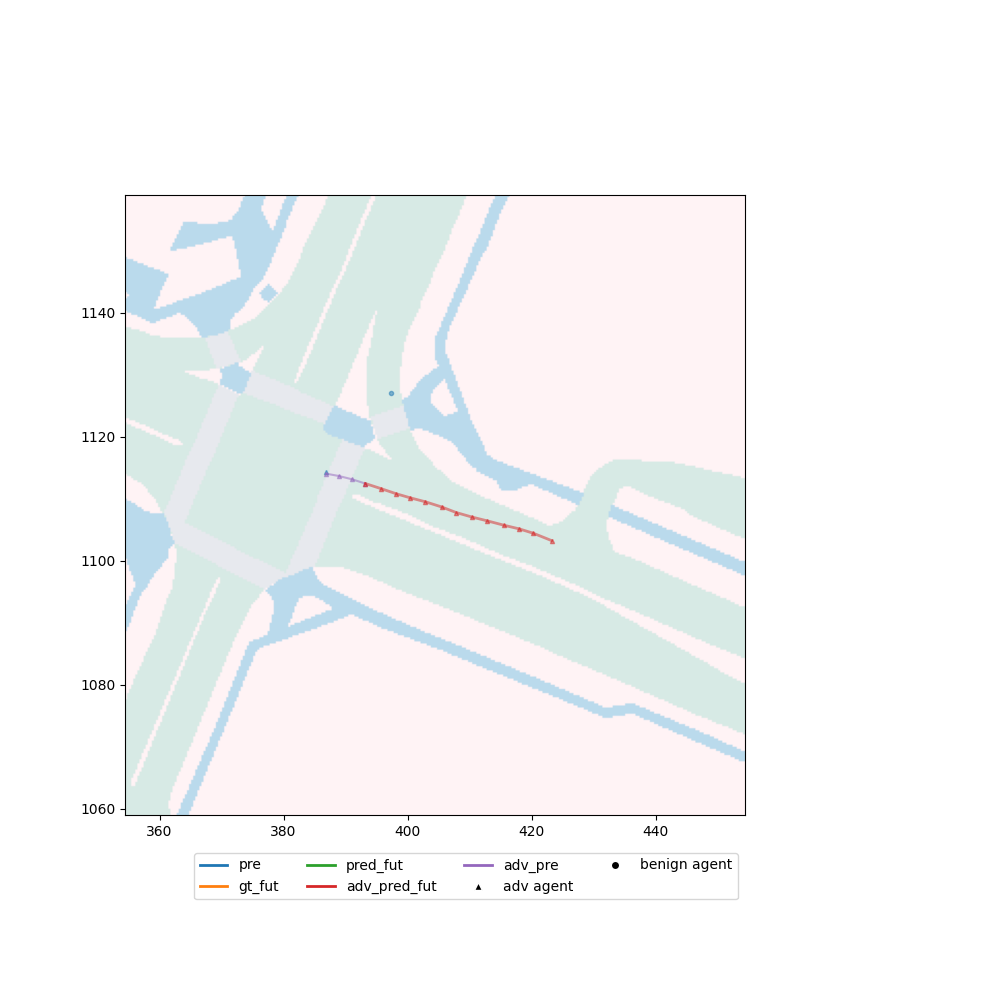} & \includegraphics[width=0.33\linewidth,clip,trim={2cm 4cm 3cm 3cm}]{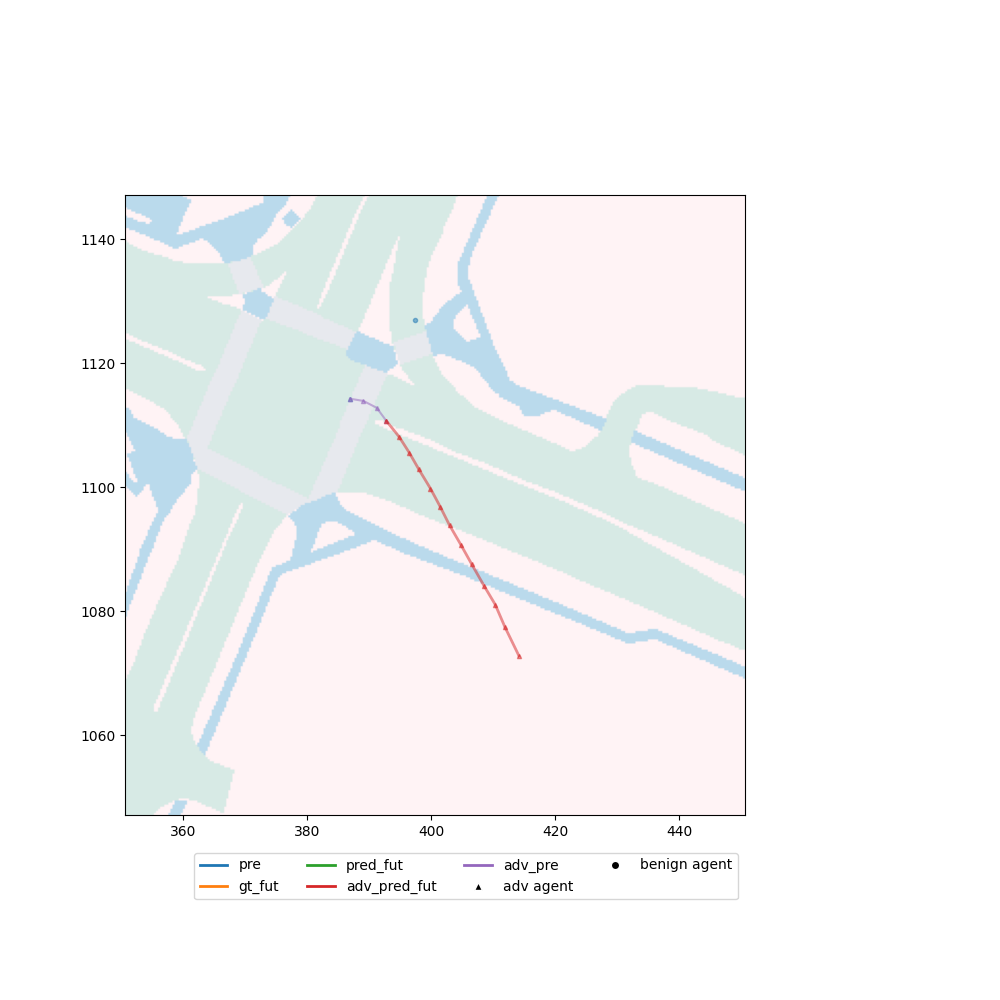} \\
  \multicolumn{3}{c}{
  \includegraphics[width=0.5\linewidth]{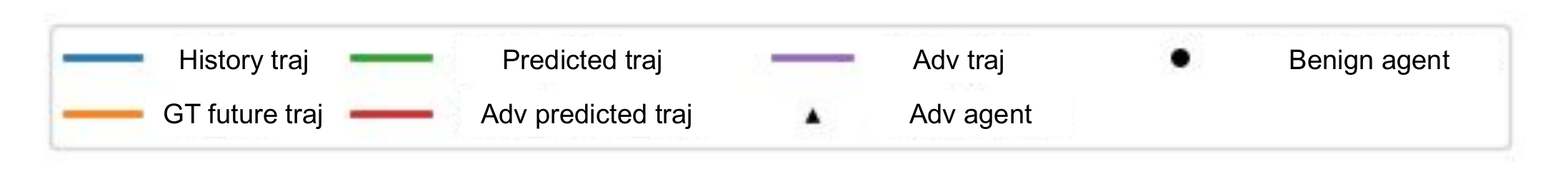}}
\end{tabular}
\end{figure}
\begin{figure}[ht]
\centering
\begin{tabular}{ccc}
 GT \& Benign Prediction & \optend{} & \searchplus{} \\
  \includegraphics[width=0.33\linewidth,clip,trim={2cm 4cm 3cm 3cm}]{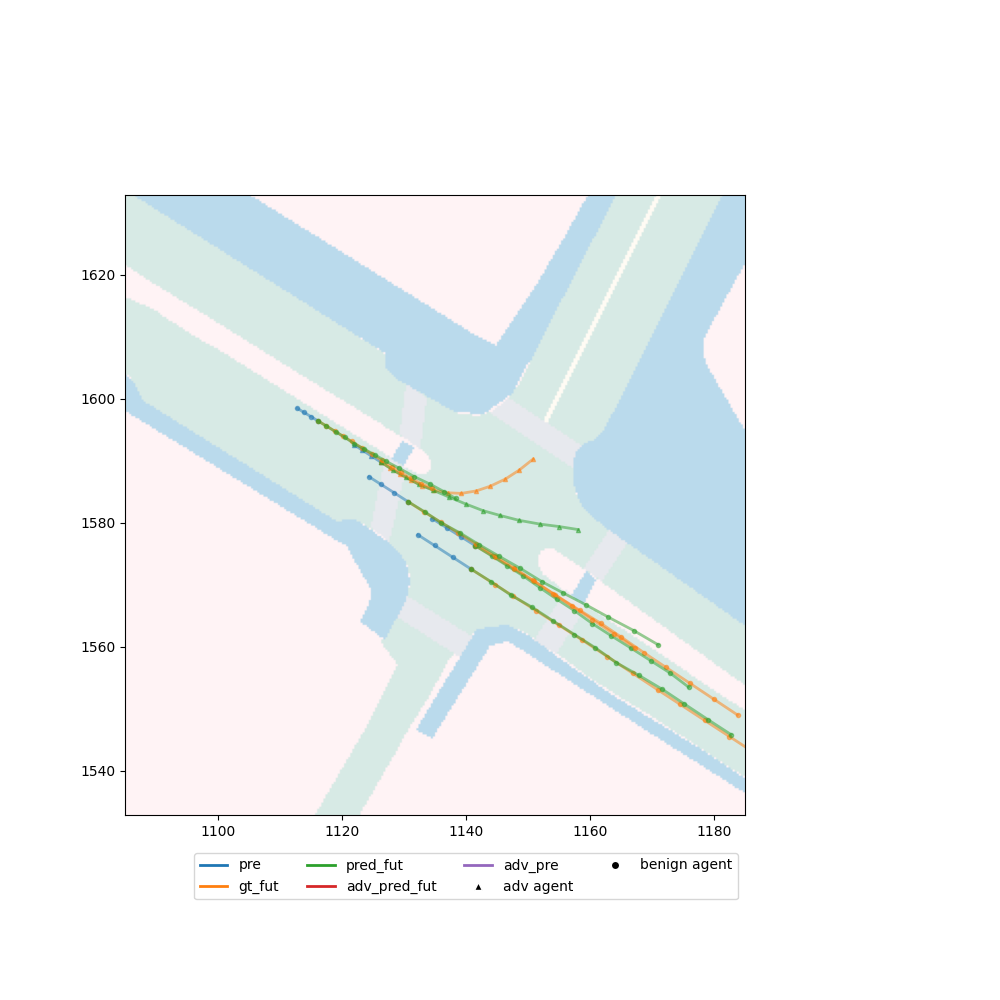} &   \includegraphics[width=0.33\linewidth,clip,trim={2cm 4cm 3cm 3cm}]{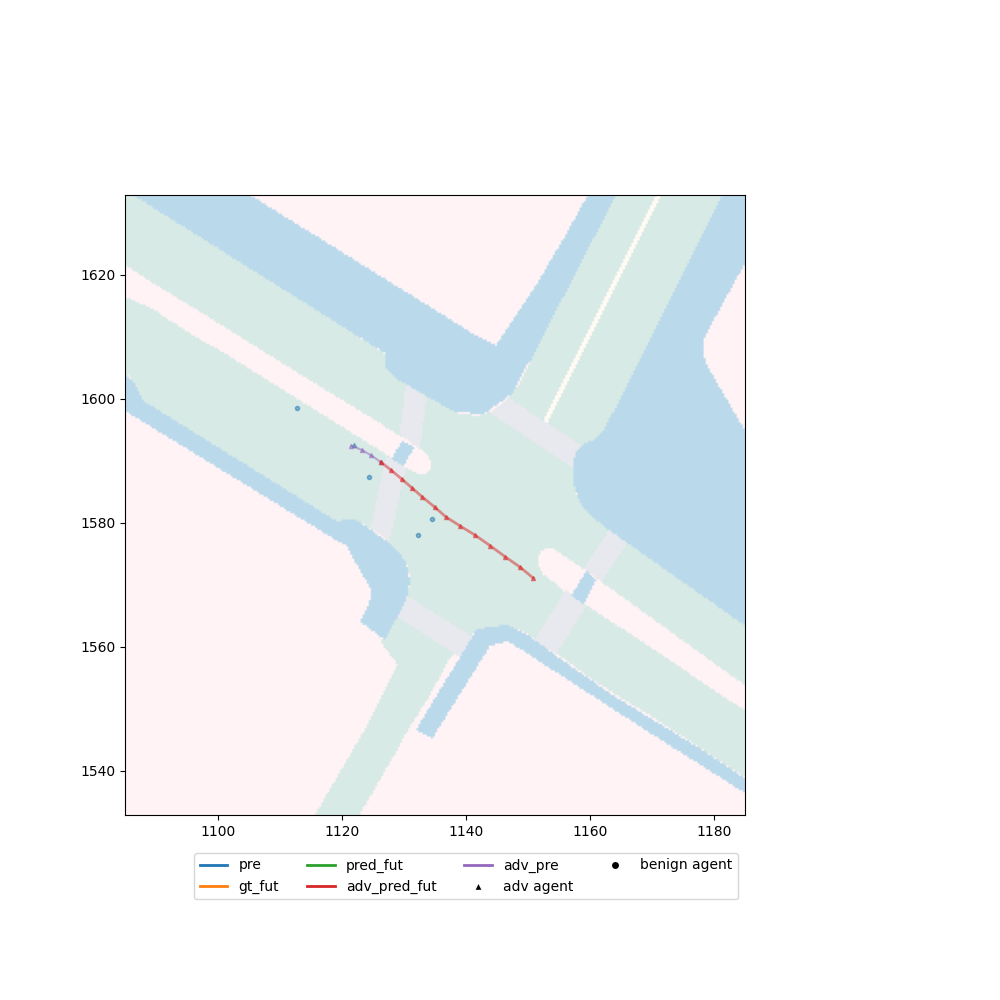} & \includegraphics[width=0.33\linewidth,clip,trim={2cm 4cm 3cm 3cm}]{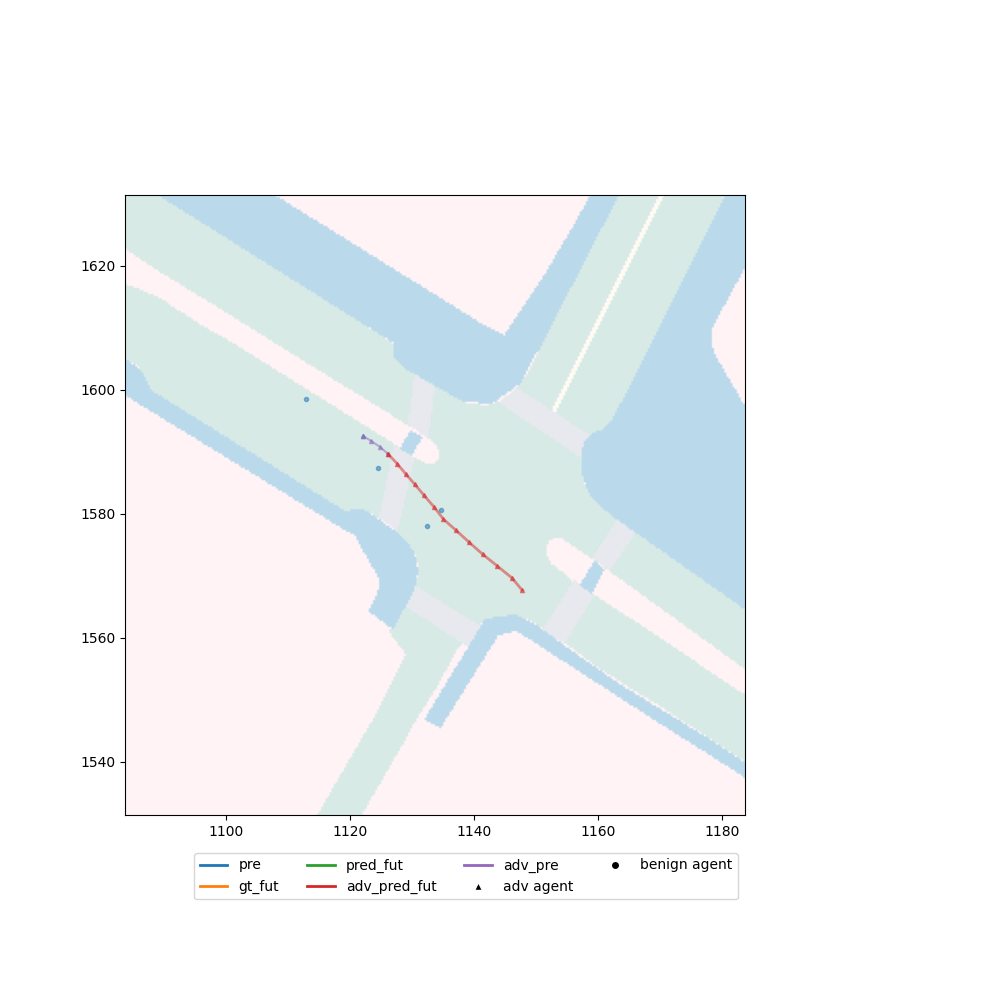} \\
  \includegraphics[width=0.33\linewidth,clip,trim={2cm 4cm 3cm 3cm}]{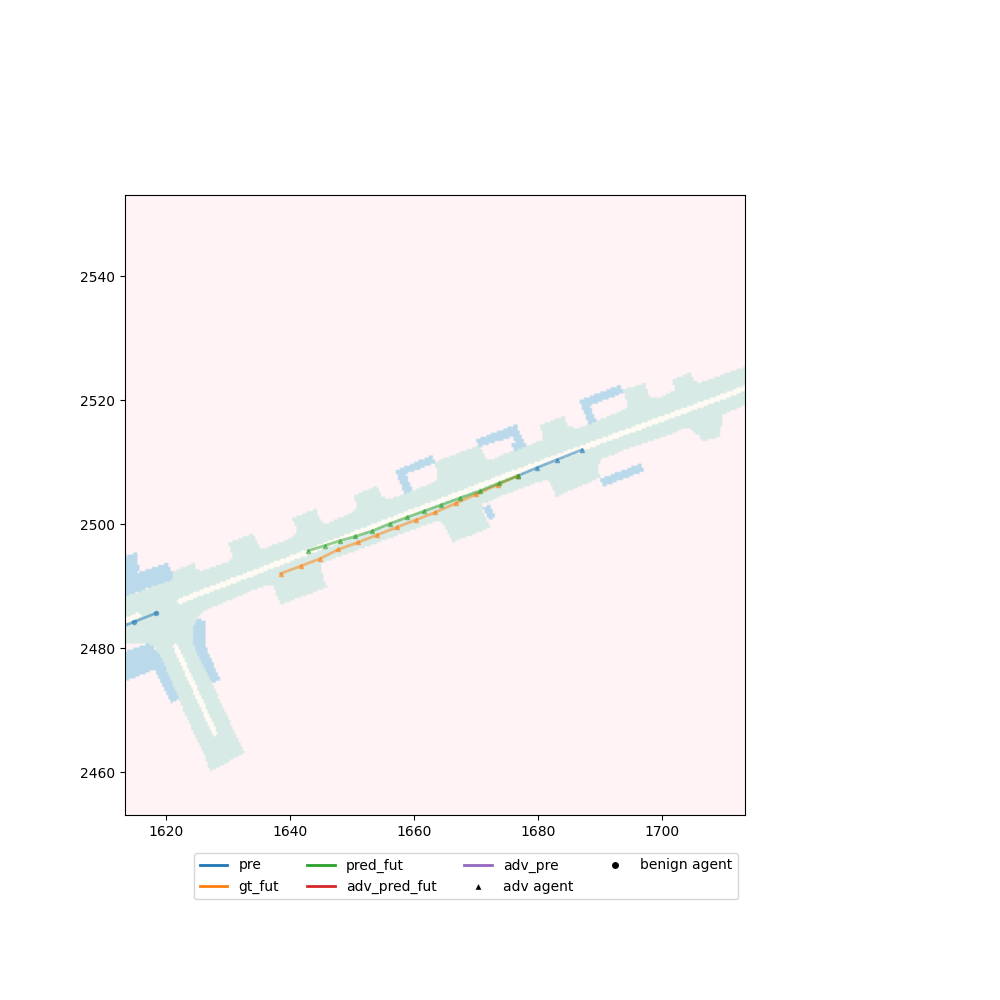} &   \includegraphics[width=0.33\linewidth,clip,trim={2cm 4cm 3cm 3cm}]{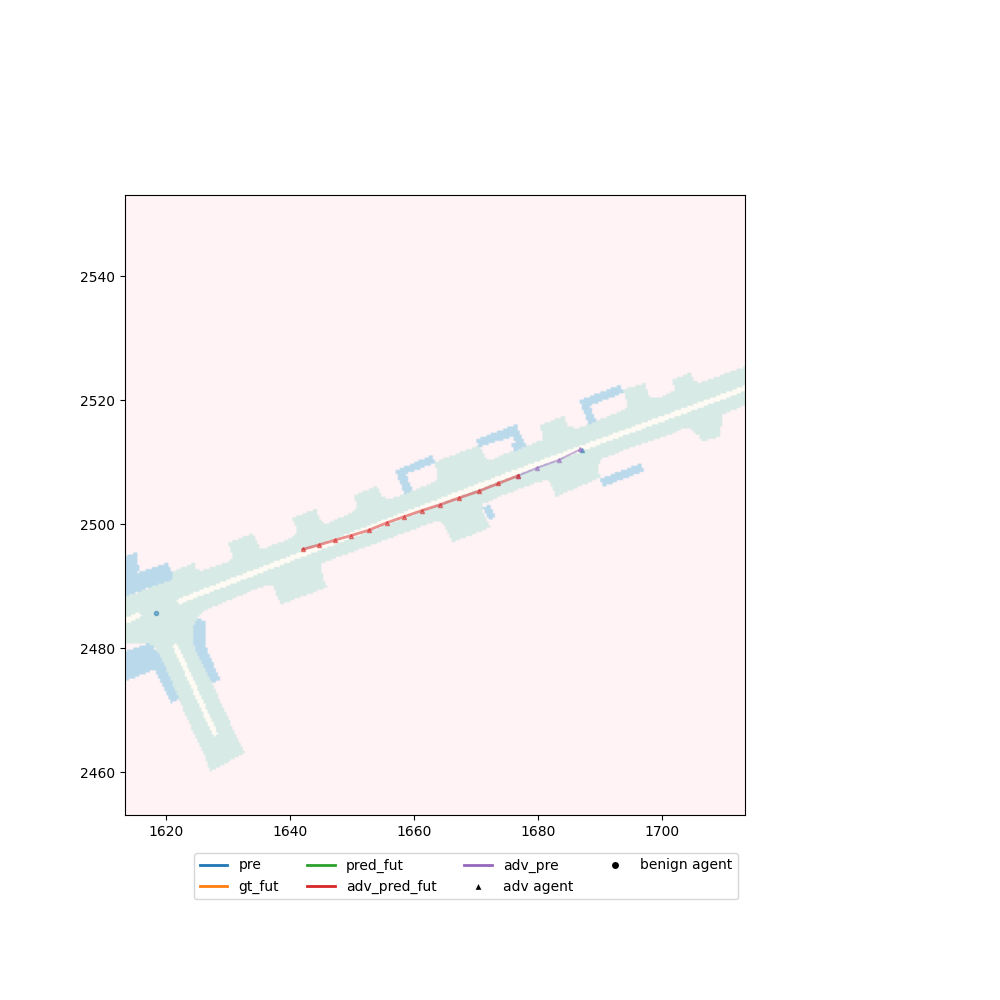} & \includegraphics[width=0.33\linewidth,clip,trim={2cm 4cm 3cm 3cm}]{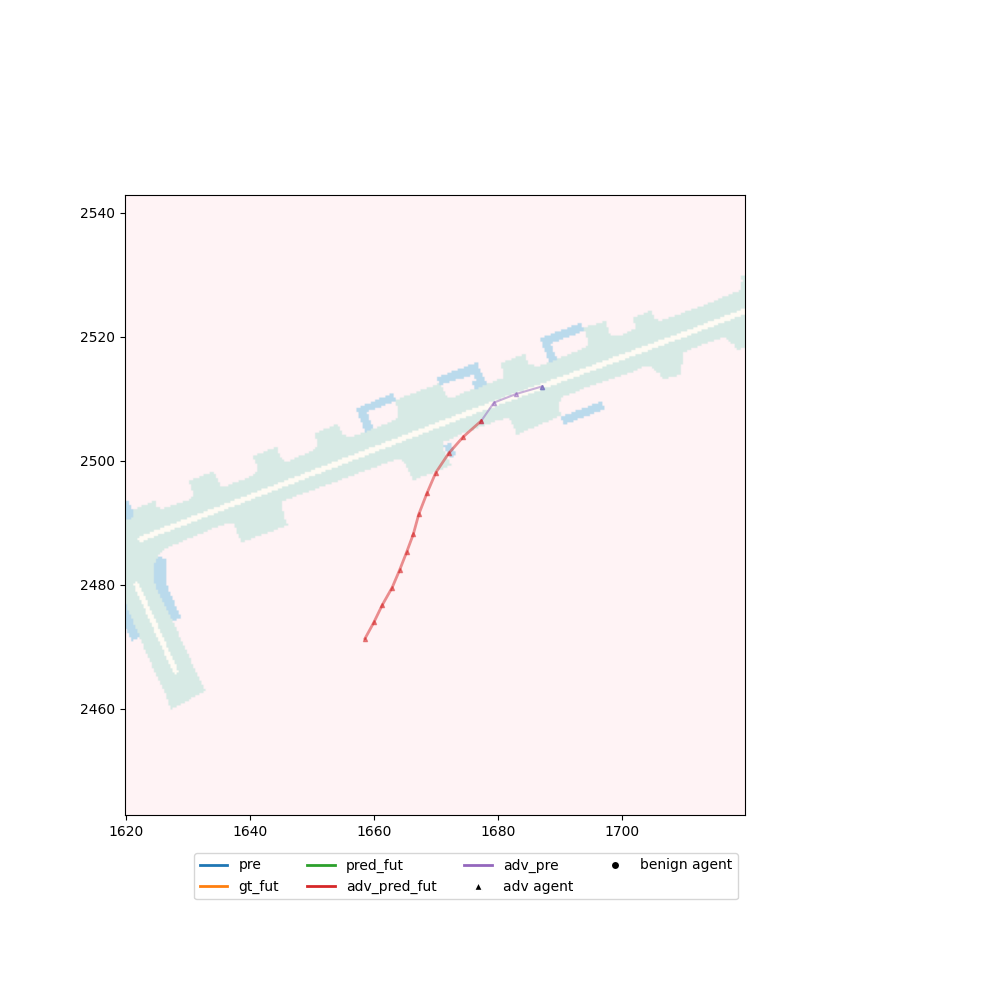} \\
  \includegraphics[width=0.33\linewidth,clip,trim={2cm 4cm 3cm 3cm}]{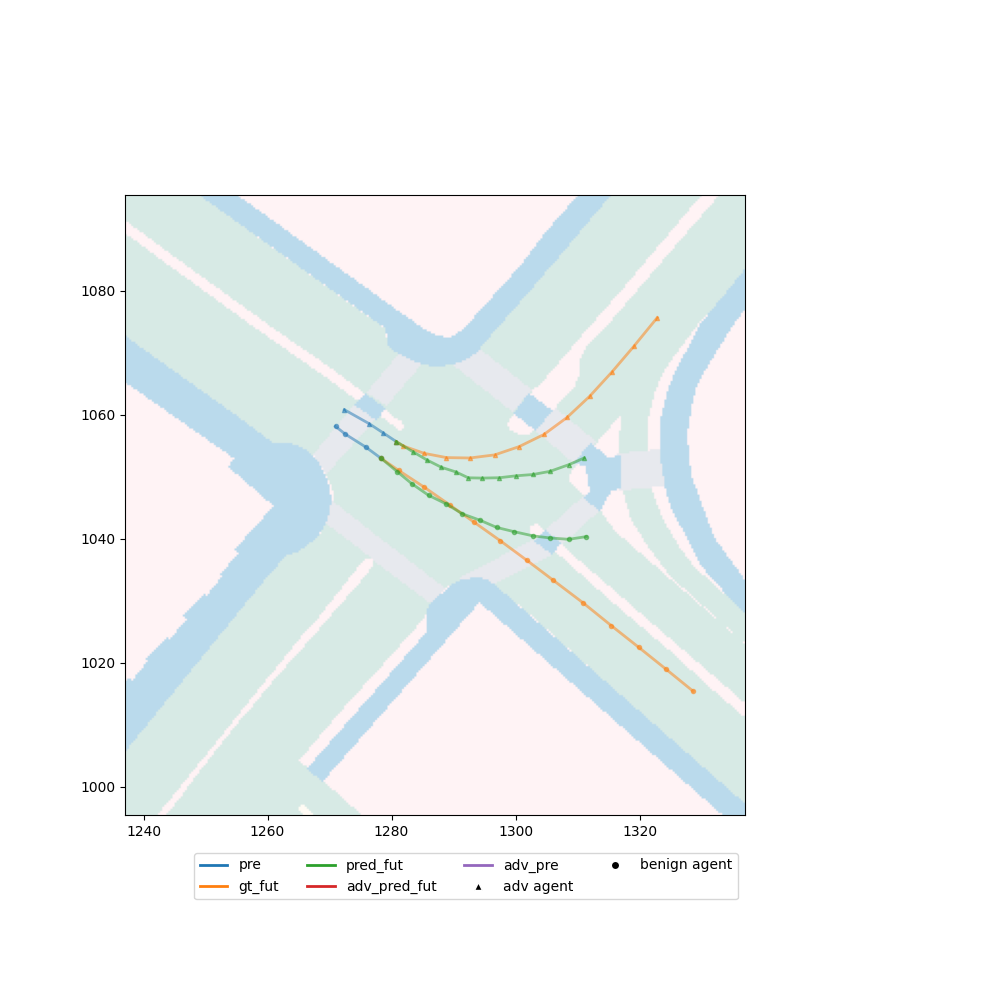} &   \includegraphics[width=0.33\linewidth,clip,trim={2cm 4cm 3cm 3cm}]{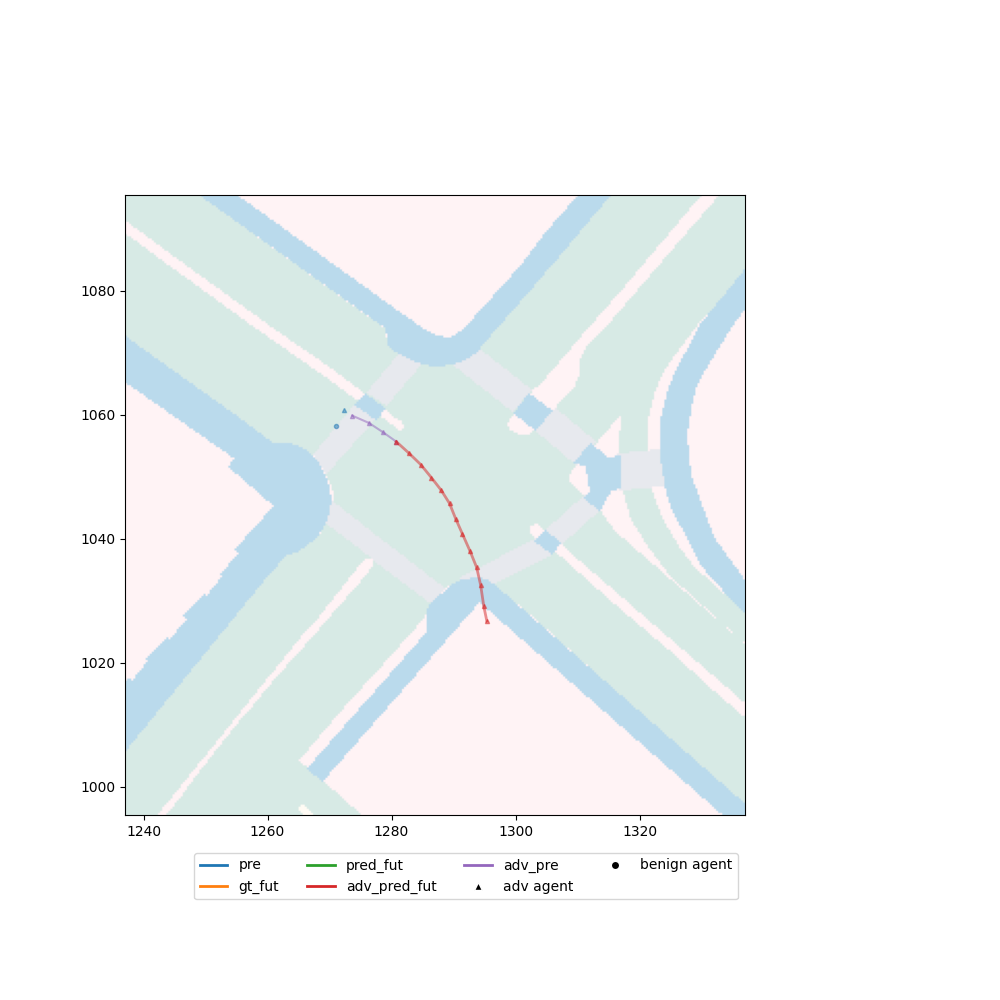} & \includegraphics[width=0.33\linewidth,clip,trim={2cm 4cm 3cm 3cm}]{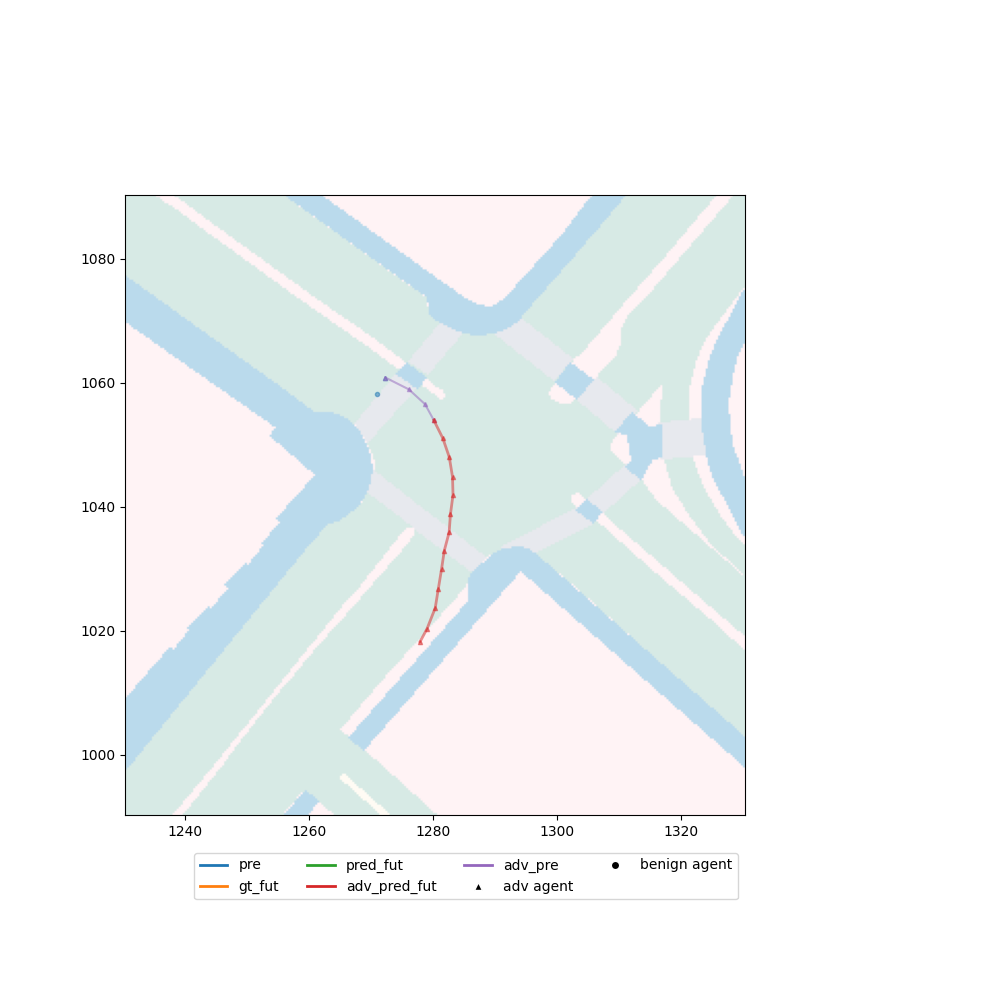} \\
  \includegraphics[width=0.33\linewidth,clip,trim={2cm 4cm 3cm 3cm}]{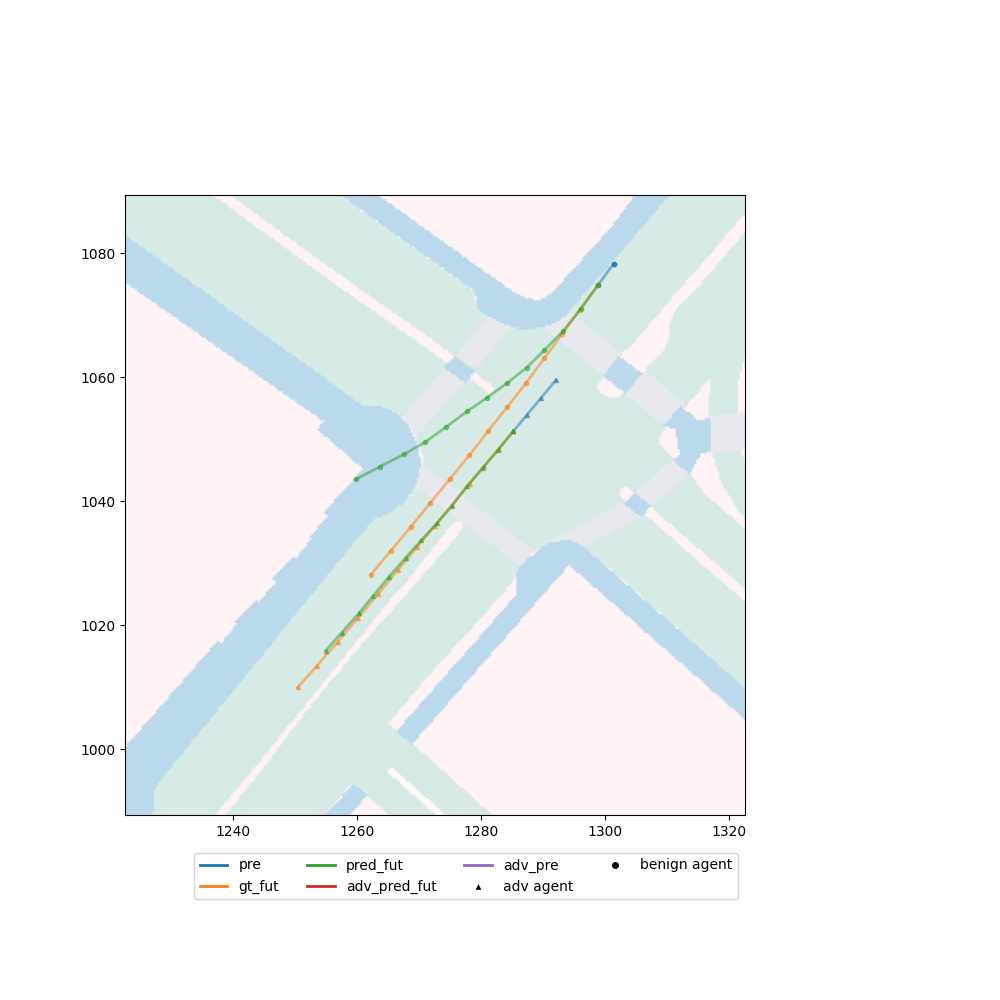} &   \includegraphics[width=0.33\linewidth,clip,trim={2cm 4cm 3cm 3cm}]{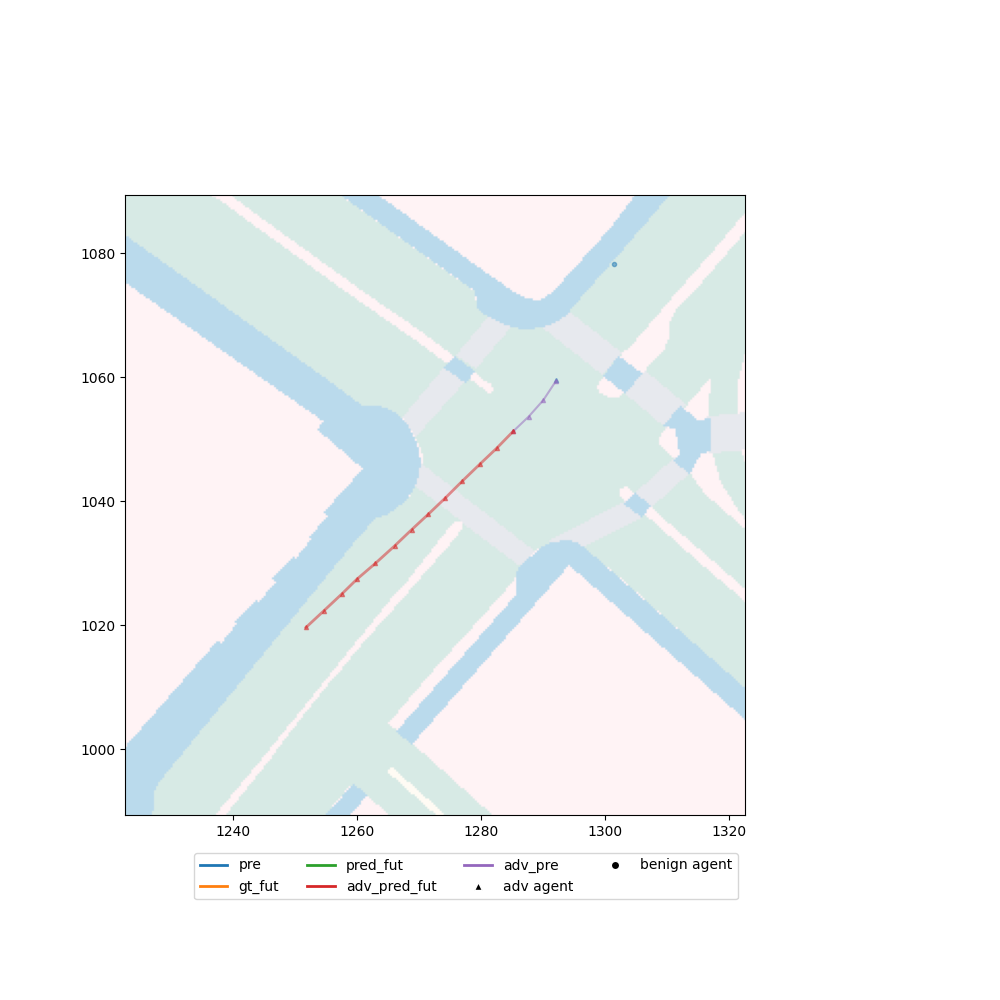} & \includegraphics[width=0.33\linewidth,clip,trim={2cm 4cm 3cm 3cm}]{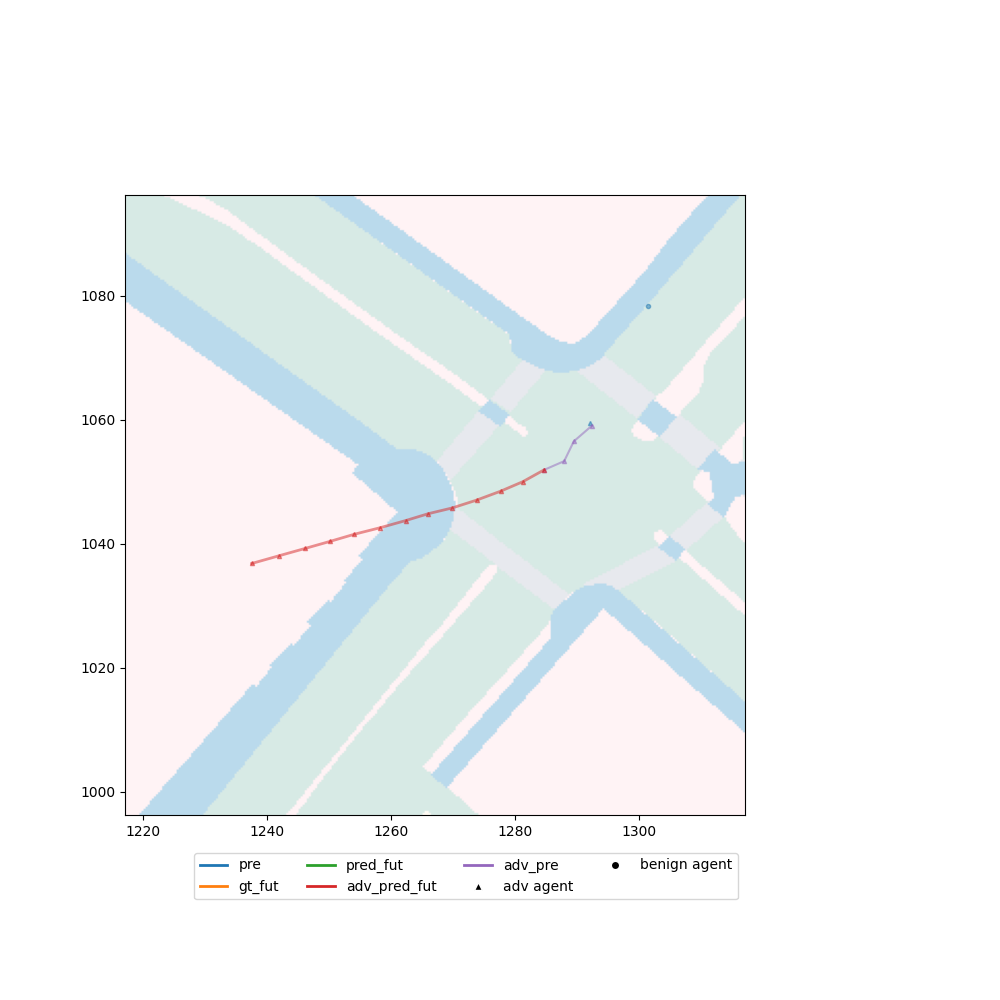} \\
  \multicolumn{3}{c}{
  \includegraphics[width=0.5\linewidth]{figs/vis_label.pdf}}
\end{tabular}
\end{figure}

\noindent\textbf{Violation rates}. Since the violation rates metric is only suitable for the \searchplus\xspace method and on the curvature $\kappa$ parameter, we represent the VR as:
$$ VR = \frac{\#\text{total adv trajectories}}{\#\text{adv trajectories violating curvature constraints}}$$.

\noindent\textbf{Aggregated sensitivity}. To approximate the behavior change quantitatively, we leverage the sensitivity concept proposed by Ivanovic et al.~\cite{Ivan2022planpred}. Sensitivity $\textbf{PI}(\textbf{Y}_i,\textbf{Y}_{\text{ego}})$ of an agent's trajectory to the ego agent represents how much the agent's trajectory $\textbf{Y}_i$ will affect the ego planning $\textbf{Y}_{\text{ego}}$. Therefore, we can present how much the adversarial trajectory $\textbf{X}_{\text{adv}}$ will affect other agents' planning $\textbf{X}_i$ as the aggregated sensitivity of the adversarial agent's trajectory to all the other agents in the scene. With a normalization over agents nearby, we attain the aggregated sensitivity:
$$ \Sigma\text{Sensitivity}(\textbf{X}_{\text{adv}},\textbf{X}) = \frac{1}{m}\sum_{i, \Vert\textbf{X}_{\text{adv}}-\textbf{X}_i\Vert < \rho}^{m} \textbf{PI}(\textbf{X}_{\text{adv}},\textbf{X}_{i})$$, where $m$ represents the total number of agents nearby filtered by the distance threshold $\rho$, which is empirically set to 5 meters. Therefore, we attain the metric for measuring behavior change as:
$$\Delta \text{Sensitivity} = \Sigma\text{Sensitivity}(\textbf{X}_{\text{adv}},\textbf{X}) - \Sigma\text{Sensitivity}(\textbf{X}_{\text{orig}},\textbf{X})$$

\noindent\textbf{Other metrics}. To measure the behavior change quantitatively, we also include evaluation results with other metrics proposed by Jekel et al. for comparing the similarity between trajectories~\cite{jekel2019}, including Dynamic Time Warping (DTW), Fréchet Distance (FD), Partial Curve Mapping (PCM), Area and Curve Length (CL). In Table~\ref{tab:behavior_other} we demonstrate that the proposed methods have lowest error for all similarity metrics. The results are also consistent with the result on $\Delta$sensitivity metric.

\begin{table}[]
\parbox{.6\linewidth}{
    \centering
    \caption{Similarity between original history trajectory and adversarial trajectory generated from \searchplus{}, \optinit{} and \optend{}.}
    \label{tab:behavior_other}
    \begin{tabular}{c|ccccc}
    \toprule
        Attack method & DTW$\downarrow$ &  FD$\downarrow$ & PCM$\downarrow$ & Area$\downarrow$ & CL$\downarrow$ \\\midrule
         \searchplus{} &  0.3558 &     0.2490  &     0.0676  &    0.8892  &  0.0003 \\
         \optinit{} & 0.2303     & 0.1429     & 0.0209    & 0.5928    & 0.0002 \\
        \optend{} & 0.1891     & 0.0564     & 0.0210     & 0.3045    & 0.0001 \\\bottomrule
    \end{tabular}}
    \hfill
    \parbox{.34\linewidth}{
    \centering
    \caption{Augmentation on AgentFormer.}
    \label{tab:aug_advdo}
    \begin{tabular}{c|c|c}
    \toprule
        & ADE & FDE  \\\midrule
        Benign & 1.83                    & 3.81                   \\
        + \textit{aug} & 1.69        &  3.57            \\
    \bottomrule
    \end{tabular}}
\end{table}

\noindent\textbf{Visualization and human study}. We randomly sample examples from 150 scenes in nuScene validation data, where the adversarial trajectory generated from \searchplus{} that have a curvature violation or a large $\Delta$Sensitivity value. In Figure~\ref{fig:adv_traj_vis}, we show that the adversarial trajectory generated from \searchplus{} have either behavior change or unrealistic steering rates. We also notice that, the \optend{} can also generate adversarial trajectory that has large turning rates but dynamically feasible. Even though the predicted results are worse under \searchplus{} attack when the curvature constraint if not bounded, \optend{} achieves higher prediction errors in average scenarios. To further show that the generated trajectories obey traffic rules, we conduct a study where adversarial trajectories are illustrated with map information (e.g. lane segments, road, crosswalk etc.). We select five human subjects with driver license and show our generated trajectories to them. Out of the 50 trajectories evaluated, only $\text{2.2}(\pm \text{1.3})$ are considered rule-violating. We conclude that the adversarial trajectory generated by our methods are more realistic in both perspectives of dynamical feasibility and behavior changing. 

\noindent\textbf{AdvDO as Augmentation.} Noticed that AdvDO also provides an opportunity for generating realistic trajectories as additional data. We replace the adversarial objective with other objectives (e.g. increasing left/right/forward/backward deviations) and generate additional data. More specifically, the objective function consists of two components:
$\mathcal{L}_\text{dyn} = \mathcal{L}_\text{d} + \gamma \mathcal{L}_\text{col}$
, where $\mathcal{L}_\text{d}$ is the deviation objective loss, $\mathcal{L}_\text{col}$ is the collision regularization loss, and $\gamma$ is a weight factor to balance the objectives. In each scene, we randomly pick a deviation objective loss $\mathcal{L}_\text{d}$ from the set $\{$moving forward, backward, left, right$\}$ for each agent. More specifically, the deviation objective loss $\mathcal{L}_\text{d}$ is formulated as 
$$ \mathcal{L}_\text{d} = (\mathbf{X} - \mathbf{X}_\text{aug}) \, \mathbf{\bar{d}} \text{,} $$
where $\mathbf{X}_\text{aug}$ represents the generated trajectories by perturbing the trajectories in the dataset and $\mathbf{\bar{d}}$ represents the unit vectors for the target deviation directions in the set of $\{$moving forward, backward, left, right$\}$. In Table~\ref{tab:aug_advdo}, we demonstrate that the augmented data improves the clean performance by 9\% on ADE. This further validates that the high fidelity of the generated trajectories with the proposed method.

\subsection{Case studies with planners}

\noindent\textbf{Planner}. In this work, to demonstrate the explicit consequences of the adversarial trajectory, we implement two planners (including path planning and motion planning). The first one is a rule-based planner as implemented by Rempe et al.~\cite{rempe2021strive}. However, we notice that this planner is enforcing path planning along the center of lane lines which leads to insufficient path sampling through the simulations. Therefore, though the planner naturally avoids driving off road, it is also lack of flexibility to dodge incoming traffic. To better represent planners equipped on AV, we implement a simple yet effective planner that uses conformal lattice~\cite{mcnaughton2011motion} for sampling paths and model predictive control (MPC)~\cite{camacho2013model} for motion planning. We call this planner MPC-based planner.

\noindent\textbf{Planning strategy.} In this work, we consider both an open-loop and a closed-loop planning strategy. Though for the closed-loop planning we have to replay the ground truth trajectories of other agents, we do notice reduced collisions and driving off road consequences and consider the closed-loop planning fashion meaningful.

\subsection{Transferability Analysis}
In this section, we aim to analyze the transferability of adversarial trajectories generated on a source model to a unseen target model. We measure the transferability by devising the \textbf{transfer rate} metric. High transfer rates indicate that the feasibility of transfer attack, which is a more realistic black-box attack, in the real-world scenario. Transfer rate is defined as the success degree of adversarial trajectories on target model over the success degree of them on source model. The success degree is measured by the average percentage of increased error (on metrics ADE/FDE/MR/ORR) with transfer attack on the target models over the increased error with white-box attack on the source models.

\subsection{Ablation Study} 
We explore the attack results in different traffic scenarios with different speeds curvatures. We calculate the aggregated speed and curvature for each agent in the entire scene to represent the speed and curvature for that scene. Similarly, we calculate the aggregated \textit{Miss Rates} to evaluate performance.

\begin{figure}[h]
  \centering
  \begin{subfigure}[b]{0.48\textwidth}
    \centering
    \includegraphics[width=\textwidth,trim={2cm 0 2cm 0},clip]{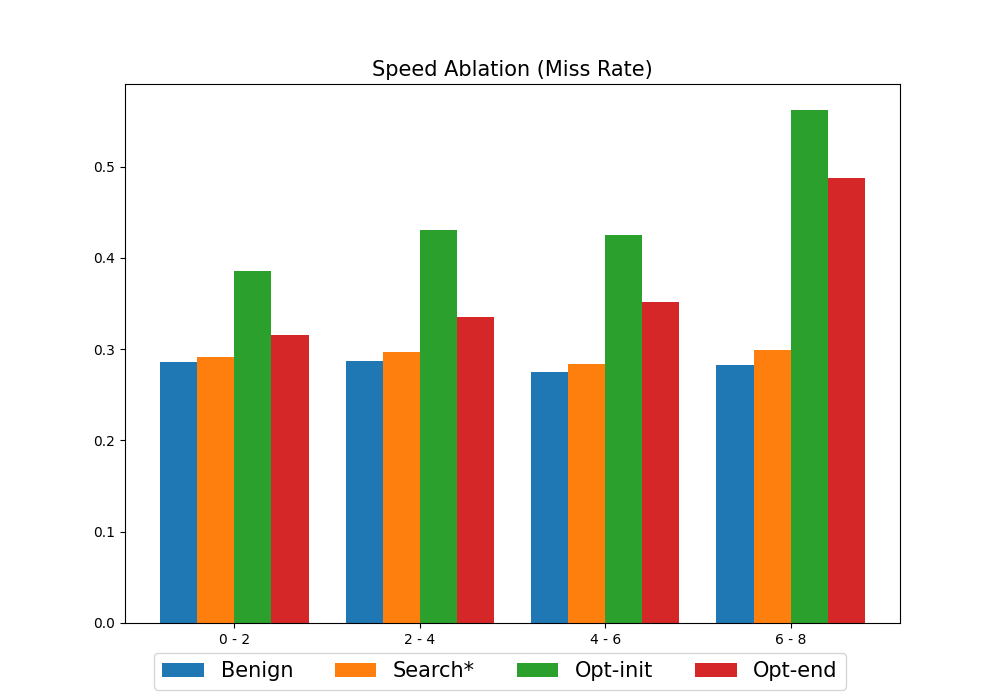}
    \caption{Speed ablation}
    \label{fig:speed_ablation}
  \end{subfigure}
  \begin{subfigure}[b]{0.48\textwidth}
  \centering
    \includegraphics[width=\textwidth,trim={2cm 0 2cm 0},clip]{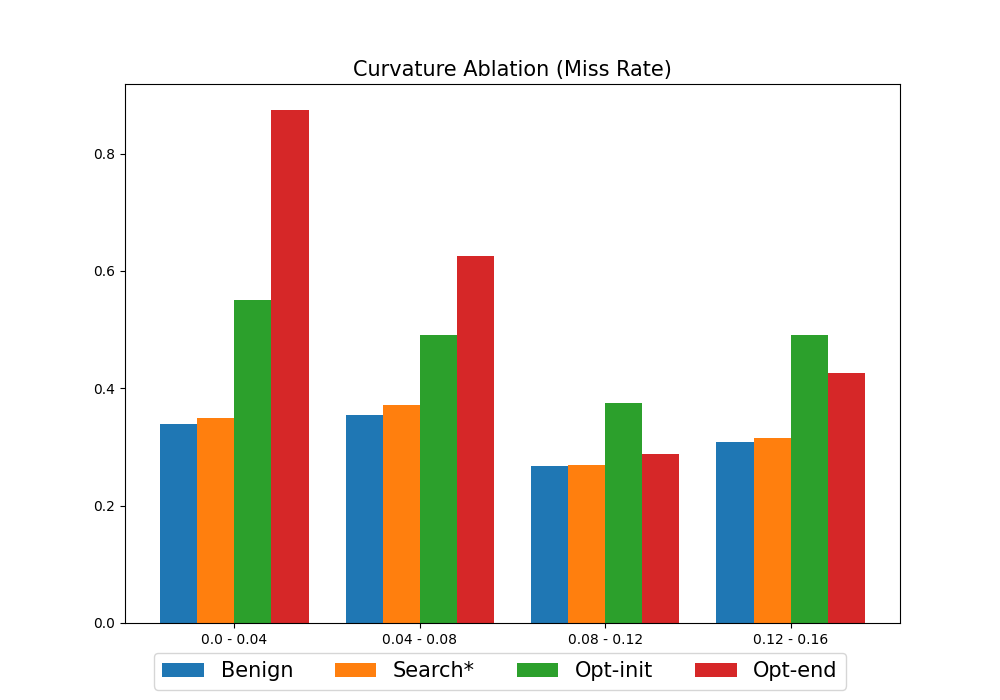}
    \caption{Curvature ablation}
    \label{fig:k_ablation}
  \end{subfigure}
  \caption{Ablation studies for different traffic scenes.}
  \label{fig:traffic_ablation}
\end{figure}

\noindent\textbf{Attack effectiveness with different speeds} 
As shown in Figure~\ref{fig:speed_ablation}, the higher speed traffic show higher \textit{Miss Rates}. It is reasonable since position deviations are larger in high speed traffics. We also notice that the attack results are consistent to results in Table 1\&2 in the main paper, which means different attack methods are not restricted due to the speed constraints.

\noindent\textbf{Attack effectiveness with different curvatures} 
In Figure~\ref{fig:k_ablation}, we notice that adversarial trajectories are more effective in small curvature traffics. This is reasonable since small curvature traffics allow more flexible adversarial trajectory generations. We find that \optend{} performs better than \optinit{} in small curvature traffic. This could be due to low curvature traffic being less sensitive to current positions.



\subsection{Mitigation}

We present a preliminary mitigation methods against adversarial trajectory via adversarial training. We notice that naive adversarial training results in noticeable degradation in benign performance for both adversarial trained models using \searchplus{} and \optinit{}. In Table~\ref{tab:adv_train_noise}, we demonstrate that the performance degradation are much smaller and even better for the adversarial trained model with proposed method \optinit{}. 

\begin{table}[h]
\begin{small}
\caption{Adversarial training results. The number in brackets represent the difference between the benign model and adversarial trained model.}
\label{tab:adv_train_noise}
\centering
\begin{tabular}{lllllc}
\toprule
Model                                                                                & Attack       & ADE 	$\downarrow$  & FDE 	$\downarrow$  & MR 	$\downarrow$     & ORR 	$\downarrow$    \\\midrule
\multirow{4}{*}{\begin{tabular}[c]{@{}l@{}}\textit{Benign}\end{tabular}} & Benign & 1.83                    & 3.81                    & 28.2\%                 & 4.7\%  \\
                                                                     & \searchplus{}     & 2.34           & 4.78          & 34.3\%      & 6.6\%        \\
                                                                     & {\optinit{}}     & {3.39}           & {5.75}           & {44.0\%}        & {10.4\%} \\\hline
\multirow{4}{*}{\begin{tabular}[c]{@{}l@{}} \textit{Rob-\searchplus}\end{tabular}} & Benign      & 2.69(+0.86)                   & 5.82(+2.01)                    & 37.8\% (+9.6\%)                & 10.2\%(+5.5\%) \\

                                            & \searchplus{}  & 2.72(+0.38)                    & 5.76(+0.98)                    & 40.7\%(+6.3\%)                 & 12.3\%(+5.8\%) \\
                                                                                     & \optinit      & 2.81(-0.58)                    & 5.92(+0.17)                    & 42.2\%(-1.8\%)                 & 13.8\%(+3.4\%) \\ \hline
\multirow{4}{*}{\begin{tabular}[c]{@{}l@{}}   \textit{Rob-ours} \end{tabular}}       & Benign       & 2.38(+0.55)                    & 5.03(+1.23)                    & 35.1\%(+6.9\%)                 & 8.1\%(+3.4\%)  \\
                                              & \searchplus{} & 2.42(+0.08)                    & 5.25(+0.47)                    & 36.8\%(+2.5\%)                 & 9.2\%(+2.6\%)  \\
                                                                                     & \optinit      & 2.54(-0.85)                    & 5.21(-0.54)                    & 36.4\%(-7.6\%)                 & 8.9\%(-1.5\%)  \\ \bottomrule
\end{tabular}
\vspace{-5pt}
\end{small}
\end{table}

\end{document}